\documentclass{article}

\usepackage{microtype}
\usepackage{graphicx}
\usepackage{subfigure}
\usepackage{booktabs} %
\usepackage[table]{xcolor}
\usepackage{algorithm}
\usepackage{algorithmic}
\usepackage{caption}
\usepackage{amsmath}
\usepackage{amssymb}
\usepackage{xcolor}
\usepackage{multirow}
\usepackage{tabularx}
\usepackage{tgcursor}
\usepackage{colortbl}

\usepackage[outline]{contour}
\usepackage[normalem]{ulem} 
\newcommand{\arrow}{\contour{black}{${\nearrow}$}}

\usepackage{tikz}

\usepackage{hyperref}

\usepackage[accepted]{icml2024}

\icmltitlerunning{Towards Modular LMs by Building and Reusing a Library of LoRAs}

\begin{document}

\renewcommand{\ttdefault}{cmtt}
\newcommand{\lora}{\texttt{lora}}

\twocolumn[
\icmltitle{Towards Modular LLMs by Building and Reusing a Library of LoRAs}
\icmlsetsymbol{equal}{*}

\begin{icmlauthorlist}
\icmlauthor{Oleksiy Ostapenko}{equal,msr,mila,udem}
\icmlauthor{Zhan Su}{equal,mila,cop} \\
\icmlauthor{Edoardo Maria Ponti}{ed}
\icmlauthor{Laurent Charlin}{mila,hec,cifar}
\icmlauthor{Nicolas Le Roux}{msr,mila,udem,cifar}
\icmlauthor{Matheus Pereira}{msr} \\
\icmlauthor{Lucas Caccia}{equal,msr}
\icmlauthor{Alessandro Sordoni}{equal,msr,mila,udem}
\end{icmlauthorlist}

\icmlaffiliation{msr}{Microsoft Research}
\icmlaffiliation{ed}{University of Edinburgh}
\icmlaffiliation{mila}{Mila --- Quebec AI Institute}
\icmlaffiliation{cop}{University of Copenhagen}
\icmlaffiliation{hec}{HEC Montréal}
\icmlaffiliation{udem}{Université de Montréal}
\icmlaffiliation{cifar}{Canada CIFAR AI Chair}

\icmlcorrespondingauthor{A.\ Sordoni}{alsordon@microsoft.com}

\icmlkeywords{Machine Learning, ICML}

\vskip 0.3in
]

\printAffiliationsAndNotice{\icmlEqualContribution} %

\begin{abstract}
The growing number of parameter-efficient adaptations of a base large language model (LLM) calls for studying whether we can reuse such trained adapters to improve performance for new tasks. We study how to best build a \emph{library} of adapters given multi-task data and devise techniques for both \emph{zero-shot} and \emph{supervised} task generalization through \emph{routing} in such library. We benchmark existing approaches to build this library and introduce model-based clustering, \texttt{MBC}, a method that groups tasks based on the similarity of their adapter parameters, indirectly optimizing for transfer across the multi-task dataset. To re-use the library, we present a novel zero-shot routing mechanism, \texttt{Arrow}, which enables dynamic selection of the most relevant adapters for new inputs without the need for retraining. We experiment with several LLMs, such as Phi-2 and Mistral, on a wide array of held-out tasks, verifying that MBC-based adapters and Arrow routing lead to superior generalization to new tasks.  We make steps towards creating modular, adaptable LLMs that can match or outperform traditional joint training.
\end{abstract}

\section{Introduction}
Tailoring large language models (LLMs) towards downstream tasks, domains, or user profiles is of paramount importance given the recent democratization of their usage, catalyzed by the release of open-source LLMs \citep[\textit{inter alia}]{zhang2023llama,microsoft_phi2}. This process often relies on an \emph{adapter}, such as LoRA~\citep{lora}, a parameter-efficient fine-tuning (PEFT) of a pre-trained LLM~\citep{lora,tfew,li-liang-2021-prefix}.

\begin{figure}[t]
\centering\includegraphics[width=0.9\columnwidth]{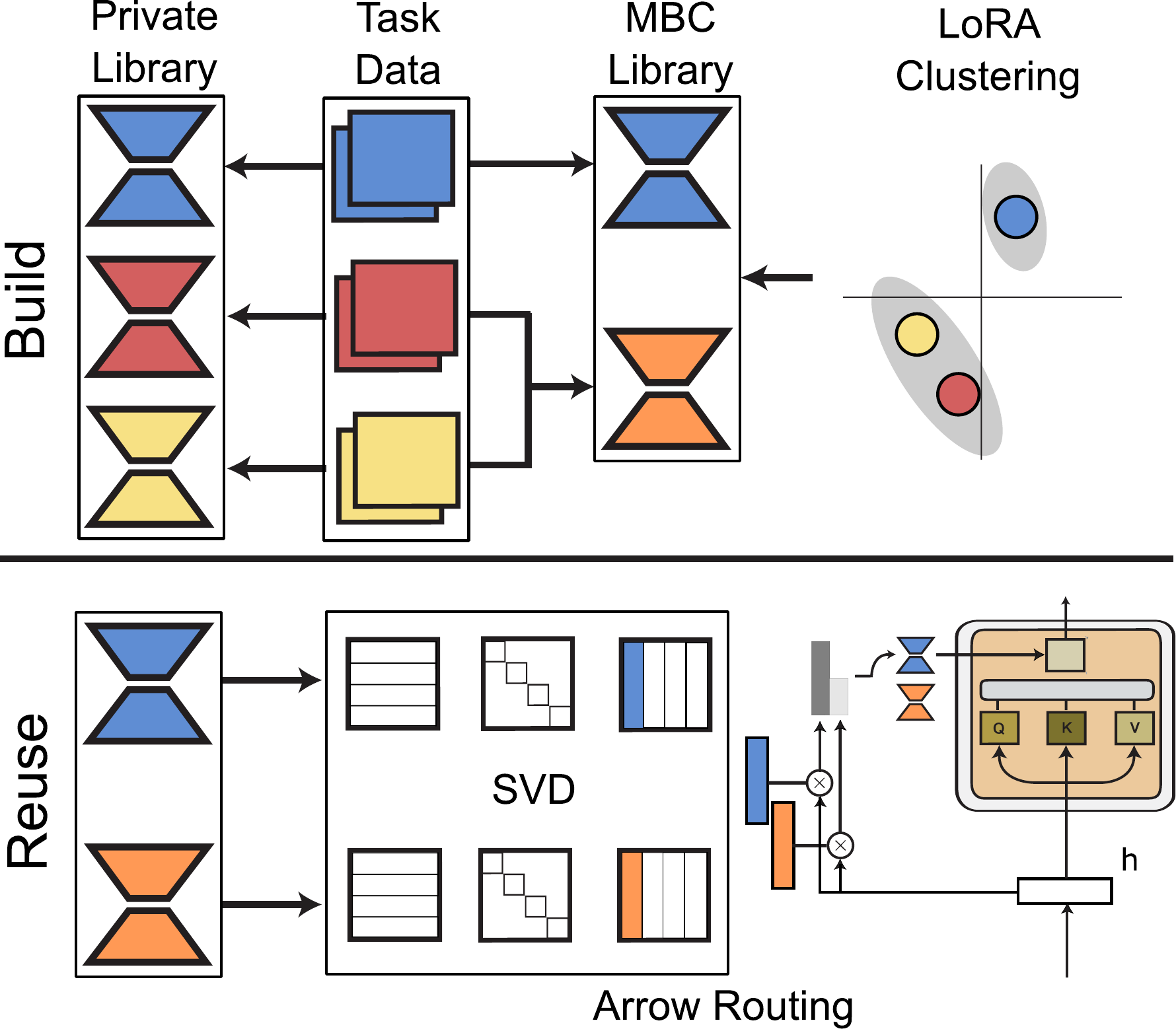}
\caption{How to coordinate a library of adapters (e.g., LoRAs) for zero-shot generalization to new tasks? To \textbf{build} this library (top), we propose \texttt{MBC}, a novel method that clusters tasks based on the similarity of the parameters of corresponding LoRAs. To \textbf{
reuse} a library (either private or \texttt{MBC}, bottom), we route hidden states to trained LoRAs via \texttt{Arrow}, which leverages the SVD decomposition of each LoRA.}
\label{fig:intro-fig}
\vspace{-5mm}
\end{figure}

LLM adapters are increasingly available as part of online hubs~\citep{beck2021adapterhub,peft}. These adapters are developed independently and asynchronously by users across the globe. Hence, they implicitly constitute a library built on top of multi-task data \citep{pfeiffer2023modular}.
Prior works show that mixtures of pretrained trained adapters can facilitate few-shot adaptation of LLMs to \emph{unseen tasks}~\citep{ponti2022combining,vu2021spot,huang2024lorahub}. Reusing pre-existing adapters in a zero-shot fashion remains less explored~\citep{jang2023exploring,belofsky2023tokenlevel}. In contrast to standard mixture-of-experts approaches~\citep{fedus2021switch}, in this setting, new inputs must be routed to independently trained experts without requiring joint training of the routing mechanism and expert parameters.

This leads to the question: how to create a modular LLM end-to-end by first building and then reusing a library of adapters for supervised adaptation and zero-shot generalization? First, given a base LLM, such as Phi-2~\citep{microsoft_phi2} or Mistral~\citep{jiang2023mistral}, we investigate building a library of adapters by leveraging 256 tasks from Flan-v2~\citep{longpre2023flan}.\footnote{We held out SNI tasks to test supervised adaptation.} We focus on LoRA~\citep{lora} and leave the extension to other adapter types for future work. Once the adapter library has been built, we devise routing strategies to evaluate~\emph{zero-shot} generalization on 10 downstream tasks comprising common-sense reasoning and coding (ARC~\citep{clark2018think}, MBPP~\cite{austin2021program}, \emph{inter alia}) and \emph{supervised adaptation} on 12 SuperNatural Instructions (SNI) tasks~\citep{wang2022super}.

\textbf{How to build the adapter library?} One straightforward approach is to operate in a \emph{private} scenario, in which one trains one adapter per task on the multi-task data and mix those adapters for unseen tasks~\citep{chronopoulou2023adaptersoup,vu2021spot,huang2024lorahub}. This method is useful when the multi-task data cannot be shared for joint training ~\citep{mireshghallah2020privacy} but trained adapters can.
To favour transfer between training tasks, recent approaches compress the multi-task data into a smaller set of reusable, composable adapters~\citep{ponti2022combining,caccia2023multihead}. In this shared data setting, we propose model-based clustering (\texttt{MBC}), a simple two-stage approach to build a library of adapters. We find a positive correlation between the similarity of the LoRA weights of a pair of tasks and the transfer between the two tasks. Building on this intuition, we first exploit LoRA similarity in weight space between privately trained adapters as a proxy for detecting clusters of similar tasks, then train one adapter per cluster. Our approach empirically improves performance while matching the compute budget.

\textbf{How to reuse the library for new scenarios?} Given a library of trained LoRAs, we examine strategies of how to reuse the library in two settings: \emph{zero-shot generalization} and parameter-efficient \emph{supervised adaptation} to new tasks.
Reusing LoRAs in a zero-shot manner is challenging because there is no labelled data to learn a routing mechanism. We propose \texttt{Arrow} (\arrow{}), a routing mechanism that automatically selects relevant LoRAs without requiring \emph{i)} joint training and \emph{ii)} access to the data used to train each LoRA. This facilitates the vision of a decentralized system where LoRAs can be trained asynchronously and be readily reused with minimal assumptions. \texttt{Arrow} computes a representation for each LoRA as the direction of maximum variance induced by the LoRA parameters. At inference time, \texttt{Arrow} routes \emph{per token} and \emph{per layer},~i.e.\ each hidden state is routed by computing its alignment with each LoRA representation. 

In summary, our contributions are: \emph{i)} we  study how to create LoRA-based modular multi-task LLM in a setting where experts are trained independently and the router is created \textit{after} the training of the experts; %
\emph{ii)} assuming shared multi-task data, we propose a clustering approach (\texttt{MBC}) to train a library of adapters; and, \emph{iii)} we propose \texttt{Arrow}, a zero-shot routing method to select which adapters to use from a library of LoRAs. This allows for routing to independently trained experts without accessing their training data.

\section{Preliminaries}
We are given a set of tasks $\mathcal{T} = \{t_1, \ldots, t_{T}\}$, where each task $t_i$ is associated with a dataset containing a set of samples  $\mathcal{D}_i = \{(\mathbf{x}_1, \mathbf{y}_1), ..., (\mathbf{x}_n, \mathbf{y}_n)\}$. The union of the training sets constitutes our multi-task dataset $\mathcal{D}$; in our case, it is Flan~\citep{longpre2023flan}.
In order to create our library of task adapters, we use LoRA~\citep{lora}. LoRA achieves competitive trade-offs between performance and parameter efficiency \citep{mahabadi2021parameter} by modifying the linear transformations in a base LM. For each linear transformation in the base LM, LoRA modifies the base model parameters as follows:
\begin{equation*}
\tag{LoRA}
h = W \mathbf{x} + s\cdot A B^\top \mathbf{x},
\label{eqn:lora}
\end{equation*}
where $W$ are the (frozen) weights of the base LM, $A, B \in \mathbb{R}^{d \times r}$ are low-rank learnable parameters and $s\ge 1$ is a tunable scalar hyperparameter. LoRA achieves parameter efficiency because of the reduced rank $r$ ($\ll d$).

\section{Building the LoRA Library}
\label{sec:library}
We propose different alternatives for building a library $\mathcal{L}$ of adapters that perform well on the tasks they were trained on and are versatile enough to be effective on other unseen downstream tasks. To do so, we seek methods that enhance multi-task transfer while reducing task interference~\citep{wang2021gradient,chen2022modsquad}.

\paragraph{\texttt{Private} Adapters}
One straightforward solution is to train separate adapters on each training task,~i.e.\ the library will be composed of $T$ adapters (see Fig.~\ref{fig:intro-fig}). Several existing methods operate in this setting, such as LoraHub~\citep{huang2024lorahub}, AdapterSoup~\citep{chronopoulou2023adaptersoup} and SPoT~\citep{vu2021spot}. Although this solution does not exploit multi-task training, 
it is required in settings where the task data is private,~e.g.,\ user data, and cannot be shared. Moreover, this setting reflects well the scenario in which adapters are trained by end users in a decentralized fashion and added asynchronously to the library.

\paragraph{\texttt{Shared} Adapter}
To encourage transfer, another solution is to train a single adapter on all the multi-task training data. One possible shortcoming is the reduced capacity to fit the multi-task training data and the possibility of interference between the multitude of training tasks~\citep{ponti2022combining}. Training a single adapter may result in negative transfer because task gradients are misaligned~\citep{wang2021gradient}. An obvious solution to reduce the amount of interference is to increase the number of trainable parameters, e.g.\ to fine-tune the whole base LM on the multi-task data~\citep{tfew}.

\paragraph{\texttt{Poly} / \texttt{MHR} Adapters}
Polytropon (Poly) and Multi-Head Routing (MHR)~\citep{ponti2022combining,caccia2023multihead} explore intermediate approaches between private and shared, where $K < T$ ``basis'' adapters are trained on the multi-task training data. These $K$ adapters can be considered ``latent skills'', as each task adapter in the multi-task training set can be expressed as a linear combination of these basis adapters. If private training for all the tasks learns a matrix of parameters $\Phi \in \mathbb{R}^{T \times D}$, where $D$ is the dimensionality of the LoRA adapters, Poly decomposes $\Phi = Z \hat\Phi$, where $Z \in \mathbb{R}^{T \times K}, \hat\Phi \in 
\mathbb{R}^{K \times D}$, $\hat\Phi$ storing the latent skills and $Z$ the linear combination coefficients for each task which specify the task-specific routing w.r.t. the latent skills. Both $Z$ and $\hat\Phi$ are trained jointly on the multi-task training set by gradient descent. Note that the skills $\hat \Phi$ do not correspond to specific tasks and therefore it is not clear how to reuse them for zero-shot generalization~\citep{caccia2023multihead}.

\begin{figure}[t]
\centering
\includegraphics[width=0.45\textwidth]{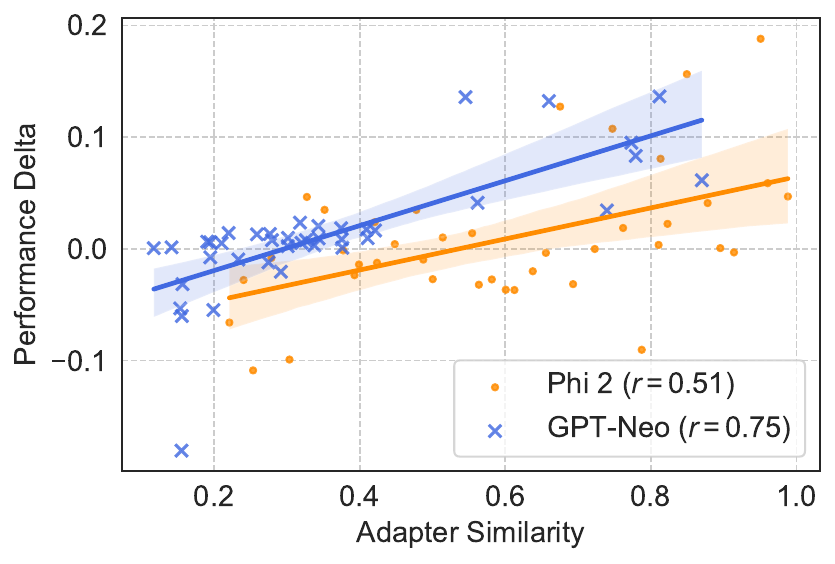}
\vspace{-4mm}
\caption{\label{fig:interf} For any pair of tasks, we report the cosine similarity between the corresponding LoRA weights (x-axis) against the delta in performance between LoRAs trained on them individually and jointly (y-axis). 
The positive correlation indicates that if LoRAs are dissimilar, we should abstain from multi-task training.}
\end{figure}

\paragraph{
Model-Based Clustering (\texttt{MBC})}
\label{sec:MBC}
While Polytropon and MHR reduce the inventory size, they require joint training of experts and the router on the combined dataset of all tasks. Here, we propose another approach to compress multi-task data into a set of reusable adapters; we cluster tasks based on their similarity and then train one adapter per task cluster. Ideally, the similarity between two tasks should correlate with the benefit of training a single model on both tasks compared to having two independent models ~\citep{fifty2021efficiently,vu-etal-2020-exploring}. 
Motivated by~\citep{zhou2022efficiently}, we rely on the intuition that LoRA parameter vectors similarity can approximate the amount of transfer between a pair of tasks. To confirm this, we devise the following experiment: we sample pairs of tasks $(t_i, t_j), t \in \mathcal{T}$ from the multi-task dataset, and we train both a) a LoRA on each task independently b) a LoRA on the union of the training datasets for the two tasks. We then compute the cosine \emph{similarity} between the flattened LoRA parameters. We quantify \emph{transfer} as the difference in the average log-likelihood induced by the joint and private models when evaluated on the test set of the two tasks. In Fig.~\ref{fig:interf}, we observe that, for two different base models (GPT-Neo and Phi-2), the closer the tasks are in LoRA parameter space, the more performance delta is when we train on the joint dataset.

\begin{algorithm}[t!]
\small
\begin{algorithmic}
   \STATE {\bfseries Input:} Multi-task data $\mathcal{D}_1, \ldots, \mathcal{D}_T$, base model {LLM}$_\theta$, number of library adapters $K$ \\
   \STATE {\bfseries Output:} Library $\mathcal{L}$ \\
   \vspace{2.1pt}
   \STATE $\mathcal{L} = \{\}$, $ \mathcal{A}= \{\}$\hfill\emph{$\triangleright$ LoRA params}
   \FOR{$t=1$ {\bfseries to} $T$}
   \STATE $A_t, B_t = \textrm{train}(\mathcal{D}_t, \textrm{LLM}_\theta)$\hfill{\emph{$\triangleright$ Train LoRA on task $t$}}
   \STATE $\mathcal{A} = \mathcal{A} \cup \{\textrm{cat}(\textrm{flatten}(A_t),\textrm{
flatten}(B_t))\}$
   \ENDFOR
   \STATE $U$ = \textrm{SVD}($\mathcal{A}$)\hfill\emph{$\triangleright$ Reduce LoRA dim} \\
   \STATE $S = \textrm{cosine-similarity($U, U$)}$\hfill\emph{$\triangleright$ $T\times T$ similarities} \\
   \STATE $c_1, \ldots, c_K = \textrm{k-means}(S, K)$ \hfill\emph{$\triangleright$ Cluster similarities}\\
   \FOR{$k=1$ {\bfseries to} $K$}
   \STATE $\mathcal{D}_k = \bigcup \mathcal{D}_t, \forall t \in c_k$\hfill\emph{$\triangleright$ Join datasets in cluster}
   \STATE $A_k, B_k = \textrm{train}(\mathcal{D}_k, \textrm{LLM}_\theta)$ \\
   \STATE $\mathcal{L} = \mathcal{L} \cup \{(A_k, B_k)\}$
   \ENDFOR \\
   \STATE \textbf{Returns} $\mathcal{L}$ \\
\end{algorithmic}
\caption{
Model-Based Clustering (\texttt{MBC})}
\label{alg:mbc}
\end{algorithm}

The previous observation warrants our simple two-stage training procedure illustrated in Fig.~\ref{fig:intro-fig} (top). Given a fixed computation training budget of $N$ training steps per task, we use the first $n$ steps to train private LoRAs. We then use these LoRA parameters to group tasks into $K$ clusters by running a standard clustering algorithm (K-Means). In the second stage of training, we train one adapter per cluster for an additional $
N - n$ training steps, which keeps the total amount of computation similar to other approaches. We refer to this method as Model-Based Clustering (\texttt{MBC}) as it uses the model-based information encoded in the weights to determine a similarity metric between tasks (see Alg.~\ref{alg:mbc}).

\section{Reusing the LoRA Library}
\label{sec:using_library}
Next, we study the reuse of a trained library $\mathcal{L}$ in two scenarios: for new inputs $\mathbf{x}^*$,~i.e.\ \emph{zero-shot}, and in a \emph{supervised adaptation} setting, where new tasks $t^*$ come equipped with their training data $\mathcal{D}_{t^*}$. While the latter has been addressed in recent works~\citep{huang2024lorahub,caccia2023multihead,vu2021spot}, the former scenario remains less explored~\citep{jang2023exploring,belofsky2023tokenlevel}.
We first devise routing strategies in the zero-shot and supervised settings and then describe how to aggregate the contributions of adapters selected by the routing strategies.

\subsection{Routing}
We denote the hidden state for any token at a given transformer layer produced by the input token $\mathbf{x}^*$ as $\mathbf{h}^*$. Similar to MoE approaches, we seek to parameterize a layer-specific routing distribution that prescribes which adapters to use. We denote this categorical distribution over $|\mathcal{L}|$ outcomes as $p(\cdot \mid \mathbf{h}^*, \mathbf{x}^*)$, where we drop the dependence on the layer for simplicity.
For example, in standard MoE approaches~\citep{fedus2021switch}, $p(\cdot \mid \mathbf{h}^*, \mathbf{x}^*) = \text{softmax}(W \mathbf{h}^*)$. Given that we relax the assumption that the routing and the library should be trained together, we must devise ways to learn such routing distribution a posteriori.

\subsubsection{Zero-Shot Routing}%
\textbf{$\mu$ Routing} One straightforward method to route to existing experts is to set the routing distribution to uniform for all layers, $p(\cdot \mid \mathbf{h}^*, \mathbf{x}^*) = [1/{|\mathcal{L}|}, \ldots, 1/{|\mathcal{L}|}]$. Despite its simplicity, $\mu$ routing was shown to be quite effective in recent work~\citep{caccia2023multihead,chronopoulou2023adaptersoup} and, due to the linearity of the LoRA adapters, effectively boils down to averaging the weights of the trained adapters uniformly.

\textbf{\texttt{TP} Routing}
Another variant treats routing as an $|\mathcal{L}|$-way classification problem. Specifically, given an input $\mathbf{x}$ belonging to task $t$ in our multi-task training set, we train a task predictor $f$ by minimizing the categorical cross-entropy loss $-\log f(\mathbf{x})[t]$, where $f(\mathbf{x})$ is a probability distribution obtained by learning a classifier on top of a T5 encoder~\citep{raffel2020exploring}. We then set $p(\cdot \mid \mathbf{h}^*, \mathbf{x}^*) = f(\mathbf{x}^*)$ at inference time. Note that the routing decisions are not dependent on the hidden state $\mathbf{h}^*$, so this is a router dependent on the whole input but independent of the particular token or layer in the Transformer. We call this predictor \texttt{TP} (Task Predictor).

\textbf{\texttt{CM} Routing} Centroid Matching (CM) computes a prototype for every expert (and for each layer) by averaging the hidden representations obtained by a forward pass of the LLM on each expert dataset. These prototypes can be stored in the columns of the routing matrix $W$. Once the prototypes for each expert have been obtained, the routing distribution is calculated by taking the cosine similarity between $\mathbf{h}^*$ and each expert prototype and finally applying softmax. This routing is similar in spirit to~\citet{jang2023exploring} and \citet{belofsky2023tokenlevel}.

\textbf{Arrow Routing \arrow{}} The rows of every routing matrix $W$ of standard MoE routing can be interpreted as expert ``prototypes''. \texttt{Arrow} prescribes a way to estimate such routing matrix in a 0-shot fashion without requiring data access. Let's denote by $\{A_i,B_i\}$ the parameters for expert $i$ at layer $\ell$, where we drop the dependency on $\ell$. The $i$-th LoRA expert transforms each token's hidden state $\mathbf{h}^*$ as $\mathbf{h}^*_i = A_i B_i^T \mathbf{h}^*$. 
\texttt{Arrow} finds a prototype for the expert $i$ by decomposing the outer product $A_i B_i^T$ with SVD and taking the right first singular vector of this transformation (see Alg.~\ref{alg:var}). The prototype determines the direction of most variance induced by expert $i$ in the space of hidden states $\mathbf{h}$. If the LoRA adapters are of rank 1,~i.e.\ $A_i, B_i \in \mathbb{D}^{D \times 1}$ the prototype for the expert $i$ will be equal to the normalized $B_i$ vector,~i.e.\ $\text{argmax}_{\mathbf{v}, \|\mathbf{v}\|_2 = 1} \|A_i B_i^T \mathbf{v}\|_2 = B_i/\|B_i\|_2$. 
In Section~\ref{sec:norm_analysis}, we provide empirical evidence that indeed, $\|A_i B_i^T \mathbf{v}\|_2$ is larger when $\mathbf{v}$ belongs to task $i$, thus motivating this routing approach. Given that both $\mathbf{v}$ and $-\mathbf{v}$ are valid singular vectors, we compute expert logits as the absolute value of the dot product between prototypes and inputs. Alg.~\ref{alg:var} details the prototype initialization and the routing step of \texttt{Arrow}. 

\texttt{Arrow} offers several advantages: a) it doesn't require access to training data; b) it routes differently in every layer and token, increasing the overall model expressivity, and c) it is compute efficient since it requires no further training and SVD decomposition can be computed efficiently for low-rank matrices~\cite{elhage2021mathematical,nakatsukasa2019low}.

\begin{algorithm}[t!]
\small
\vspace{1pt}
\begin{algorithmic}
   \STATE { \hspace{-12pt}  \bfseries \underline{Weight Initialization}}\\
   \vspace{1pt}
   \STATE {\bfseries Input:} LoRA library $\mathcal{L}$, layer $\ell$\\
   \STATE {\bfseries Output:} Routing parameters for layer $\ell$: $W_\ell$
   \vspace{2.5pt}
   \FOR{$i=1$ {\bfseries to} $L$}
   \STATE $A_i, B_i = \mathcal{L}[i, \ell]$\;\;\;\hfill\emph{$\triangleright$ Get weights for expert $i$}
   \STATE $U, D, V = \textrm{SVD}(A_i B_i^T)$
   \STATE $W_\ell
[i] = V[:, 0]$\hfill\emph{$\triangleright$ First right singular vector}
   \ENDFOR
   \STATE \textbf{Returns} $W_\ell$
\end{algorithmic}
\vspace{3pt}
\begin{algorithmic}
   \STATE { \hspace{-12pt}  \bfseries  \underline{Routing}} \\
   \vspace{2.5pt}
   \STATE {\bfseries Input:} Routing parameters for layer $\ell$: $W_\ell \in \mathbb{R}^{|\mathcal{L}| \times d}$, token in \\ \hspace{25pt} layer $\ell$:  $\mathbf{h}_\ell \in \mathbb{R}^{d}$, top-k routing: $k$
   \STATE {\bfseries Output:} Routing probabilities for layer $\ell$: $\mathbf{p}_\ell$
   \vspace{1.2pt}
   \STATE \text{logits} = abs($W_\ell \mathbf{h}_\ell$)
   \STATE $p_{\ell}[i] = \begin{cases}
       \textrm{logits}[i] & \textrm{if} \; i \in \textrm{arg top-}k(\textrm{logits}) \\
       -\infty & \textrm{else}
   \end{cases}$
   \STATE \textbf{Returns} $\textrm{softmax}(\mathbf{p}_\ell)$
\end{algorithmic}
\caption{\label{alg:var}\texttt{Arrow} Routing \arrow{}}
\end{algorithm}

\subsubsection{Supervised Task Routing}
When generalizing to a new task, we can learn the optimal routing given the task data $\mathcal{D}^*$. This setting is similar to previous task generalization works~\citep{ponti2022combining,caccia2023multihead,huang2024lorahub}. We compare results in this supervised setting to both Poly~\citep{ponti2022combining} and LoraHub~\citep{huang2024lorahub}.

\textbf{\texttt{Poly} Routing} treats the distribution over experts at each layer as an $|\mathcal{L}|$-dimensional parameter that is learned by minimizing the cross-entropy on the new task data $\mathcal{D}^*$. It optimizes the merging coefficients of LoRAs for the new task, i.e. $A^* = \sum_{i = 1}^{|\mathcal{L}|} w^i A_{i}$ and $B^* = \sum_{i = 1}^{|\mathcal{L}|} w^i B_{i}$. Here $p(\cdot | \mathbf{h}^*, \mathbf{x}) = (w^1, \ldots, w^n)$ is the (input-independent) learnable routing distribution for a given layer.

\textbf{\texttt{LoraHub} Routing} \citep{huang2024lorahub} is similar to \texttt{Poly} with the exception that a) it resorts to gradient-free optimization to learn routing coefficients and b) it doesn't fine-tune the experts' parameters, making it less expressive than \texttt{Poly}.

\textbf{\texttt{$\pi$-tuning} Routing} uses Fisher Information to create an embedding for each task-specific expert. In the fine-tuning process, $\pi$-tuning first trains an expert for the next task, then it retrieves a subset of experts most similar to the target task's expert using FIM embeddings. Finally, both the interpolation coefficients and experts' parameters are tuned on the target task's data \citep{wu2023pi}. 

\subsection{LoRA Composition}
Given a routing distribution $\mathbf{w} = p(\cdot \mid \mathbf{h}^*, \mathbf{x})$ obtained either using the previously presented zero-shot or supervised routing, we linearly combine adapters in the library, i.e.\ $A^* = \sum_{i=1}^{|\mathcal{L}|} w_i A_i$, $B^* = \sum_{i=1}^{|\mathcal{L}|} w_i B_i$ and use the resulting adapter to perform inference at every layer of the base LLM~\citep{ponti2022combining,huang2024lorahub}. For 0-shot task generalization, we employ top-$k$ routing, composing the $k$ experts with the highest routing logits.

\section{Experiments}
Our experimental evaluation aims to answer the following questions: 1) How does building a LoRA library compare to non-modular methods (e.g.\ full fine-tuning)? 2) How large is the gap between privately trained libraries (similar to online hubs) and libraries which assume access to multi-task data? 3) To what extent does routing facilitate reusing a library of LoRA adapters?

\label{sec:experiments}
\textbf{Multi-Task Dataset} We train expert modules on 256 tasks from the original Flan v2 dataset \cite{longpre2023flan}. We exclude the SNI tasks ($>1000$ tasks) \cite{wang2022super} from training for computational reasons. %
We reserved 12 SNI tasks for downstream out-of-domain evaluation. %
Similarly to \citet{wang2023far}, we sub-sampled 10,000 examples per task to ensure computational feasibility. Within these samples, 1,000 are allocated for validation and early-stopping. We will release our dataset for reproducibility.

\textbf{Evaluation} For our supervised adaptation study, we use 12 held-out SNI tasks, each corresponding to a different SNI category. We threshold the number of training examples to 10,000 examples per task. We evaluate performance with Rouge-L scores~\cite{lin2003automatic}. For zero-shot evaluation, we mainly use ten tasks widespread in the literature, including 1) common-sense reasoning: WinoGrande~\cite{sakaguchi2021winogrande}, HellaSwag~\cite{zellers2019hellaswag}, PIQA~\cite{bisk2020piqa}; 2) question answering: boolQ~\cite{clark2019boolq}, OpenbookQA~\cite{mihaylov2018can}, ARC-easy and hard~\cite{clark2018think}; 3) coding: (HumanEval~\cite{chen2021evaluating}, MBPP~\cite{austin2021program}; 4) general-purpose reasoning: BBH.~\citep{suzgun2022challenging}\footnote{We test on a subset of 1000 randomly sampled examples to reduce evaluation costs.} We remove overlaps between the evaluation tasks and the Flan multi-task training set (boolQ, ARC, WinoGrande, HellaSwag, OpenbookQA and PIQA). We also include zero-shot results on the 12 held-out SNI tasks in the appendix.%

\begin{table*}[t!]
\small
\centering
\begin{tabular}{llcc|cccccccccc|c} 
\toprule
&Library & Route & $|\mathcal{L}|$ & \textsc{piqa} & \textsc{boolq} & \textsc{wg} & \textsc{hswag} & \textsc{arcE} & \textsc{arcC} & \textsc{HE} & \textsc{oqa} & \textsc{bbh} & \textsc{mbpp} & Acc. \\
\midrule
\midrule
\multirow{9}{*}{\rotatebox[origin=c]{90}{\emph{\textbf{Phi-2 (2.8B)}}}} &\texttt{Base} & - & -& 79.2 & 82.7 & 75.7 & 72.5 & 77.5 & 52.9 & 45.1 & 49.8 & 48.0 & 56.0 & 63.8 \\
&\texttt{FullFT}  & - & - & 80.3 & 80.8 & 77.0 & 73.2 & 83.5 & 57.9 & 50.0 & 48.0 & 47.7 & 57.2 & 65.6 \\
&\texttt{Shared}         & - & 1& 80.4 & 82.4 & 76.6 & 73.4 & 83.2 & 55.8 & 46.3 & 50.4 & 48.4 & 58.4 & 65.5 \\
&\texttt{Poly}    & $\mu$ & 8 &80.6 & 82.3 & 76.7 & 71.7 & 82.7 & 55.3 & 48.2 & 50.4 & 49.8 & 59.1 & 65.7 \\
&\texttt{MHR}      & $\mu$ & 8 & 80.1 & 83.0 & 77.1 & 70.4 & 83.2 & 55.5 & 46.3 & 53.4 & 52.0 &  58.0 & 65.9 \\
\cmidrule(lr){2-15}
& \texttt{Private} & $\mu$ & 256& 79.5&83.2&76.0&73.1&81.4&54.5&43.9&47.8&48.5&59.9&64.8 \\
& \cellcolor{lightgray!20}\texttt{Private} & \cellcolor{lightgray!20}\arrow{}& \cellcolor{lightgray!20}256 & \cellcolor{lightgray!20}80.2& \cellcolor{lightgray!20}84.3&\cellcolor{lightgray!20}77.6& \cellcolor{lightgray!20}72.6&\cellcolor{lightgray!20}84.2&\cellcolor{lightgray!20}56.4& \cellcolor{lightgray!20}50.6& \cellcolor{lightgray!20}52.2& \cellcolor{lightgray!20}47.7& \cellcolor{lightgray!20}59.9& \cellcolor{lightgray!20}\underline{66.6} \\
\cmidrule(lr){2-15}
&\texttt{MBC} & $\mu$ & 10 &80.3 & 85.1 & 77.3 & 73.1 & 84.3 & 57.7 & 48.8 & 50.2 & 51.6 & 62.3& 67.1\\
& \cellcolor{lightgray!20}\texttt{MBC} & \cellcolor{lightgray!20}\arrow{} &\cellcolor{lightgray!20}10&\cellcolor{lightgray!20}79.9& \cellcolor{lightgray!20}84.7&\cellcolor{lightgray!20}77.7& \cellcolor{lightgray!20}72.9& \cellcolor{lightgray!20}84.8& \cellcolor{lightgray!20}57.9 & \cellcolor{lightgray!20}51.8 & \cellcolor{lightgray!20}50.2& \cellcolor{lightgray!20}52.2& \cellcolor{lightgray!20}62.3&  \cellcolor{lightgray!20}\underline{67.4}\\
\midrule
\midrule
\multirow{6}{*}{\rotatebox[origin=c]{90}{\emph{\textbf{Mistral (7B) }}}} & \texttt{Base} & - & - & 81.1  & 82.2 & 66.5  & 78.8  & 68.9 & 49.6 & 28.0 & 44.6 & 47.9 & 47.5 & 59.5 \\ 
&\texttt{Shared} & - & 1& 50.4 & 84.6 & 68.6 & 79.5 & 84.8 & 60.0 & 24.4 & 50.4 & 49.2 & 47.5 & 63.1 \\
\cmidrule(lr){2-15}
& \texttt{Private} & $\mu$ &  256 & 82.1 & 82.7 & 67.2 & 79.6 & 78.7 & 54.8 & 29.9 & 45.2 & 49.0 & 49.4 &  61.9 \\
&\cellcolor{lightgray!20}\texttt{Private} & \cellcolor{lightgray!20}\arrow{} & \cellcolor{lightgray!20}256 & \cellcolor{lightgray!20}82.8& \cellcolor{lightgray!20}86.6& \cellcolor{lightgray!20}66.6& \cellcolor{lightgray!20}81.1& \cellcolor{lightgray!20}85.7& \cellcolor{lightgray!20}60.8& \cellcolor{lightgray!20}30.5& \cellcolor{lightgray!20}50.6& \cellcolor{lightgray!20}49.5& \cellcolor{lightgray!20}49.4& \cellcolor{lightgray!20}\underline{64.4} \\
\cmidrule(lr){2-15}
& \texttt{MBC} & $\mu$ &10& 83.0 & 87.6 & 68.5 & 80.8 & 86.2 & 60.9 & 28.7 & 48.6 & 51.5 & 50.2 & \underline{64.6} \\
\rowcolor{lightgray!20}\cellcolor{white}& \texttt{MBC} & \arrow{} &10& 82.8& 87.3& 70.6& 80.9& 84.5& 59.6& 28.0& 52.8& 45.5& 47.1& 63.9 \\
\bottomrule
\end{tabular}
\caption{Downstream zero-shot results for Phi-2 and Mistral backbones. $|\mathcal{L}|$ denotes the library size. For comparison with other routing baselines, see Fig.~\ref{fig:routing_ablation}.}
\label{tab:zeroshot_full}
\end{table*}

\begin{figure}[t!]
\centering
\includegraphics[width=0.9\columnwidth]{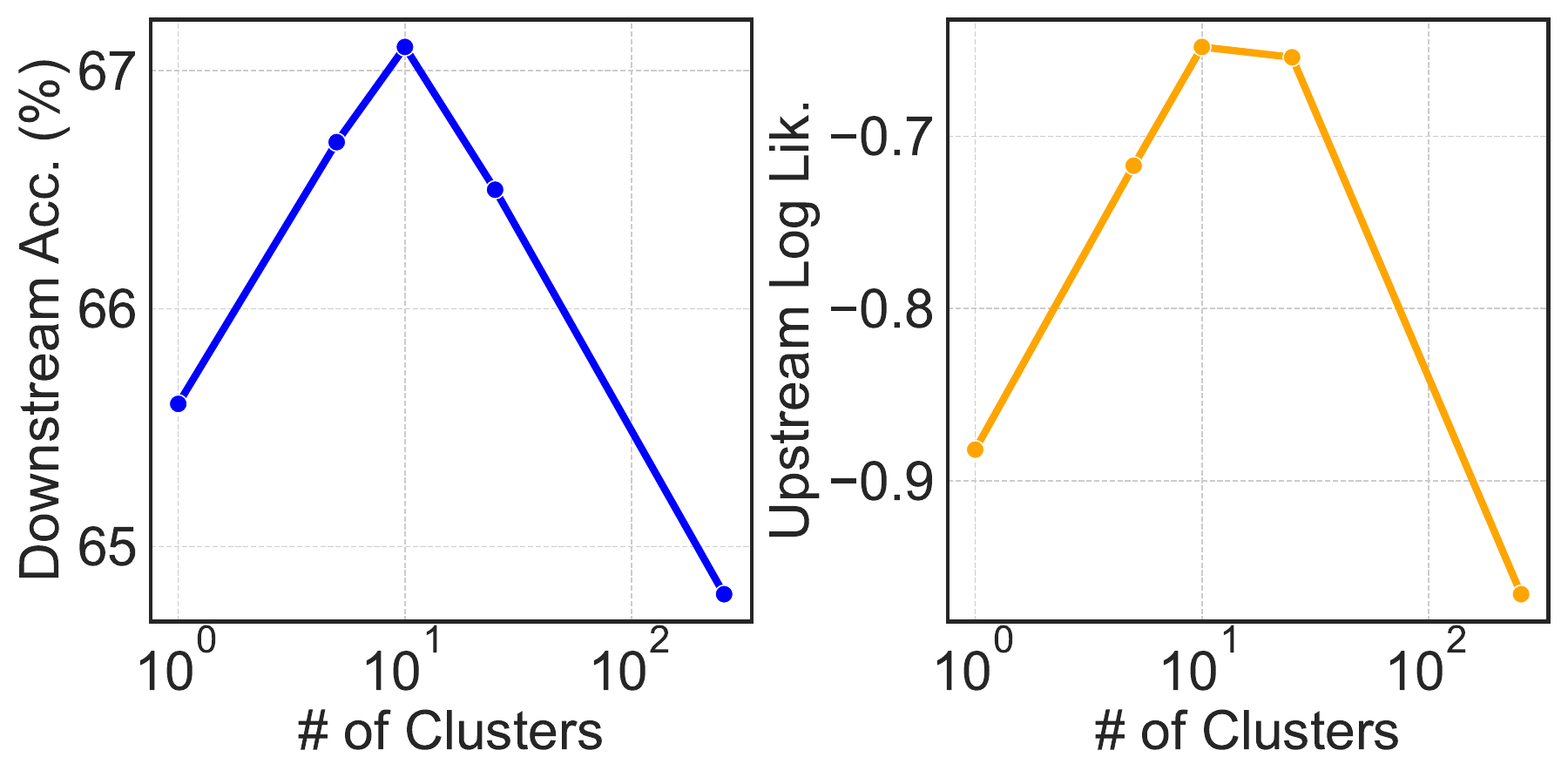}
\vspace{-3mm}
\caption{Phi-2 zero-shot accuracy on the 10 held-out tasks (\emph{left}) and validation log-likelihood on the training tasks (\emph{right}) as a function of the number of \texttt{MBC} clusters.}
\label{fig:cluster_ablation}
\end{figure}

\textbf{Models and Training} This work focuses on augmenting LLMs with a library of adapters to transform them into modular architectures. Our primary focus is on Phi-2~\cite{microsoft_phi2}, a state-of-the-art model (as of March 2024) with 2.8 billion parameters, leading its class of models with parameter counts below 3 billion, according to the open leaderboard~\cite{open-llm-leaderboard}. Additionally, we conducted experiments using the larger Mistral 7B~\cite{jiang2023mistral} model, given its widespread use in the community. For all models, we only patch attention layers with LoRA adapters. Unless stated otherwise, for all our multi-task training and single-task adaptation scenarios, we use LoRA rank of 4, dropout of 0.05 and learning rate of 1e-4. Unless specified, we set the number of clusters for \texttt{MBC} to 10, resulting in the best upstream validation loss and downstream performance for Phi-2, as demonstrated in Fig.~\ref{fig:cluster_ablation}.

\textbf{Methods} We consider the following methods in both zero-shot and supervised scenarios (except for \texttt{FullFT}): 
\begin{itemize}
    \item \texttt{Base}: the base model tested without any adaptation;
    \item \texttt{Shared}: a single expert LoRA finetuned on the joint training set of all tasks (256 tasks unless stated otherwise) on top of the base model with multi-task learning;
    \item \texttt{FullFT}: like \texttt{Shared} but the full model is finetuned.
\end{itemize}
We adopt the following naming convention for the models using a library of experts: $<$\texttt{library}$>$--$<$\texttt{routing}$>$. For the library type, we consider \texttt{Poly}, \texttt{MHR},  \texttt{Private} and \texttt{MBC} libraries described in Sec.~\ref{sec:library}. For \texttt{MBC}, we match the total amount of compute, meaning that we use 40\% of the training steps to compute the LoRA clustering and the other 60\% to compute the final cluster adapters. For routing, we use $\mu$, \texttt{TP}, \texttt{CM} and \texttt{Arrow} in the zero-shot scenario and \texttt{Poly} and \texttt{LoraHub}\footnote{For \texttt{LoraHub}, we match the amount of compute used by SGD. Assuming the backward pass is twice the compute of a forward pass, and since nevergrad~\citep[NG;][]{nevergrad} only does forward passes, to match the compute of 5 SGD training epochs, we perform 30 epochs of NG with 1/2 of the training data used by SGD methods.} for the supervised scenario, described in Sec.~\ref{sec:using_library}.

\begin{figure*}[t]
\centering
\includegraphics[width=\textwidth]{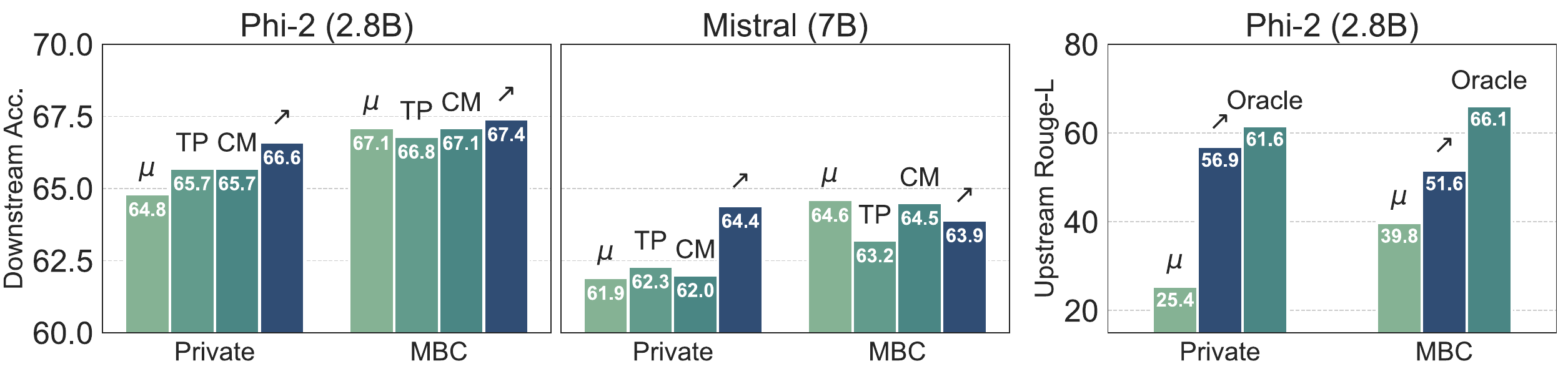}
\vspace{-4mm}
\caption{Comparison of routing approaches with both \texttt{Private} and \texttt
{MBC} libraries. \emph{\underline{Left \& Middle.}} Downstream zero-shot performance on two backbones; \texttt{Arrow} outperforms other routing approaches in the case of private libraries, while in the case of \texttt{MBC} libraries, routing is less important. \emph{\underline{Right.}}  Upstream performance on the held-out sets of each of the 256 training tasks. \texttt{Arrow} nearly matches Oracle routing (which uses information about the task identity) in the case of \texttt{Private} library and noticeably improves for \texttt{MBC}.}
\label{fig:routing_ablation}
\end{figure*}

\subsection{Zero-Shot Results}
In the zero-shot scenario, downstream tasks are evaluated without further fine-tuning. Tab.~\ref{tab:zeroshot_full} presents the mean downstream accuracy for 10 held-out tasks. First, we analyze Phi-2 results. We observe that \texttt{MHR}-$\mu$ achieves strong zero-shot performance, competitive with \texttt{Shared} and \texttt{FullFT}, in line with the results of~\citet{caccia2023multihead}. Interestingly, training one adapter per task and then taking the average, \texttt{Private}-$\mu$, still achieves gains w.r.t.\ \texttt{Base}, albeit falling short of multi-task training (\texttt{FullFT} and \texttt{Shared}), highlighting the competitiveness of uniform ($\mu$) adapter routing~\citep{chronopoulou2023adaptersoup}. Comparing the performance of our proposed \texttt{MBC} approach for library construction (\texttt{MBC}-$\mu$) to previous approaches, we notice a sizable bump in performance of 1.2\% absolute accuracy points over the strongest baseline (\texttt{MHR}). Similarly, when studying the zero-shot performance of Phi-2 on 12 SNI tasks in Tab.~\ref{tab:zero_shot_sni_phi2} we observe that \texttt{MBC}-$\mu$ strongly outperforms other baselines. Importantly, both \texttt{Shared} and \texttt{FullFT} methods, as well as \texttt{Poly} and \texttt{MHR} libraries assume simultaneous access to the full dataset of all tasks. In contrast,~\texttt{Private} and \texttt{MBC} libraries can be trained in an embarrassingly parallel manner and therefore do not require any distributed training infrastructure~\cite{li2022branch}.

Next, we analyze whether more informed routing strategies can improve performance beyond the $\mu-$routing. The full results are reported in Figure~\ref{fig:routing_ablation} (\emph{Left \& Middle}). We see that $\texttt{TP}$, $\texttt{CM}$ and $\texttt{Arrow}$ routing improve the performance over $\mu$ routing for the \texttt{Private} Phi-2 library, gaining 0.9\%, 0.9\% and 1.8\% points respectively.
This highlights the importance of routing for larger libraries. Notably, \texttt{Arrow} (66.6\%) can surpass the performance of \texttt{FullFT} (65.5\%) when applied to the \texttt{Private} library.

On the \texttt{MBC} library, $\texttt{TP}$ routing decreases performance when compared to uniform routing, while \texttt{MBC-}\arrow{} improves over \texttt{MBC}-$\mu$ by 0.3\% points and proves itself as a more robust routing method for both \texttt{Private} and \texttt{MBC} libraries. Overall, \texttt{MBC-}\arrow{} improves 3.6 points over the base model and 1.8\% absolute over \texttt{FullFT}.

For Mistral, we find a similar trend with \texttt{MBC} libraries achieving the best performance. \texttt{Arrow} routing results in a 2.5\% increase in average performance over $\mu$ routing when used with the \texttt{Private} library (\texttt{Private-}\arrow{} vs. \texttt{Private-}$\mu$). \texttt{Arrow} is able to narrow  the performance gap with~\texttt{MBC}, without requiring simultaneous data access across tasks. We do not see any gains from using other routing methods for 10 experts in the \texttt{MBC} library in this case. We make similar observations analyzing 0-shot SNI-12 results presented in Table~\ref{tab:zero_shot_sni_phi2}, where \texttt{Private-}\arrow{} attains notable gains of 10 Rouge-L points over \texttt{Private-}$\mu$ while \texttt{MBC-$\mu$} strongly outperforms all other baselines.

\textbf{\texttt{MBC} Analysis} Overall, \texttt{MBC} enhances the performance of the library across all our results. To investigate this further, we compare different clustering techniques. First, we compare to clusters obtained by randomly selecting examples (\texttt{RandomExamples}). This is equivalent to randomly partitioning the joint multi-task dataset. Then, we compare to clusters obtained by randomly choosing tasks from the entire set of training tasks ({\texttt{RandomTask}). Finally, we cluster task embeddings, which are obtained by forwarding task-specific examples through the model and averaging their representation at the model's penultimate layer (\texttt{Embeddings}). For all these methods, we set the number of clusters to 10. 

The results are shown in Table \ref{tab:cluster_ablation}.
\texttt{RandomTask} surpasses \texttt{RandomExamples} by 1.6\%, which indicates that grouping tasks rather than task examples is crucial for positive transfer.
\texttt{Embeddings} underperforms \texttt{MBC} and supports our observation that the cosine similarity between the weights of privately-trained LoRA correlates better than using representation similarity for 0-shot generalization. Additionally, we also report average pairwise cluster ``similarity'' (as measured by the cosine similarity of the LoRA weights for each cluster) and observe a tendency that expert clusters with lower similarity, i.e. higher diversity, tend to result in higher performance. We conjecture that this stems from different clusters contributing distinct features to the joint model; however, we leave further investigation in this direction to future work~\citep{jolicoeurmartineau2023population}.

\subsection{Upstream Performance}
We further assess the efficacy of~\texttt{Arrow} routing by looking at the \emph{upstream} in-distribution performance, measured as the average of the Rouge-L on the validation sets of the 256 training tasks. Within this setting, we can compute the performance of the~\texttt{Oracle} routing, which selects for each task the corresponding expert. In Fig.~\ref{fig:routing_ablation} (\emph{Right}) we report the results for \texttt{Arrow} and $\mu$ routing with both \texttt{MBC} and \texttt{Private} libraries.
For both libraries, $\arrow{}$ increases performance w.r.t. $\mu$ and almost matches~\texttt{Oracle} performance in the \texttt{Private} setting. This demonstrates \texttt{Arrow}'s ability to correctly select the most relevant modules from a large library of experts.

\begin{table}[t]
\centering
\small
\begin{tabular}{lc|c}
\toprule
Clustering & Mean Acc. & Similarity\\
\midrule
\midrule
\texttt{RandExamples$^*$-$\mu$} & 64.8 & 0.82 \\
\texttt{RandTask$^*$-$\mu$}  & 66.4 & 0.58 \\
\texttt{RandTask-$\mu$}  & 66.4 & 0.58 \\
\texttt{Embeddings$^*$-$\mu$} & 66.1 & 0.37 \\
\texttt{MBC$^*$-$\mu$} & 66.7 & 0.37\\
\texttt{MBC-$\mu$} & 67.1 & 0.27 \\
\bottomrule 
\end{tabular}
\caption{Ablation of task clustering: \texttt{RandTask} clusters \textit{tasks} randomly, \texttt{RandExamples} clusters \textit{examples} randomly, \texttt{Embeddings} clusters examples based on their embedding similarity. `*' denotes one epoch of training to save computation. We also report average cosine similarity between cluster adapters.}
\label{tab:cluster_ablation}
\end{table}

\subsection{Supervised Adaptation} 
In Table~\ref{tab:supervised_adaptation}, we present the supervised adaptation results for Phi-2 on the full (100\% of training data) and limited (10\% of training data) data regimes. The detailed per-task performance as well as the adaptation results for the Mistral model are presented in Table~\ref{tab:adaptation_results_10} and \ref{tab:adaptation_results_100}. First, for all models (Phi-2, Mistral) we observe a notable performance boost coming from using \texttt{Private} and \texttt{MBC} libraries compared to \texttt{No Library}, which optimizes a LoRA for each downstream task by starting from a random initialization, and~\texttt{Shared}, which starts from the multi-task trained LoRA solution. Secondly, similarly to zero-shot results, we observe that \texttt{MBC} can boost the performance with both \texttt{Poly} and $\mu$ routing: for Phi-2 the performance of \texttt{MBC}-$\mu$ tops \texttt{Private}-$\mu$. Additionally, we see that randomly grouping tasks \texttt{RandomTask-Poly} outperforms the non-library baselines but does not quite match \texttt{MBC}-based clustering for all the models.
The low performance of \texttt{LoraHub} can be attributed to the fact that \texttt{LoraHub} does not fine-tune the LoRA experts' weights but only their routing coefficients (due to gradient free optimization). Refer to App.~\ref{sec:data_scarse} for more insights onto this point. Finally, \texttt{MBC}-$\mu$ performs similarly to \texttt{MBC-Poly}, echoing results in~\citep{caccia2023multihead}.

\begin{table}[t]
\centering
\small
\begin{tabular}{lcccc}
\toprule
Method & 100\% Data & 10\% Data\\
\midrule
\midrule
\texttt{Base} & 22.2 & 22.2 \\
\texttt{No Library} & 75.5 & 53.9 \\
\texttt{Shared} & 75.8 & 56.4 \\
\texttt{Poly} & 73.4 & 61.7 \\
\texttt{MHR} & 74.8 & 64.5 \\
\texttt{$\pi$-tuning} & 76.7  &  64.6 \\
\midrule
\texttt{Private-$\mu$} &
 76.9  & 62.5 \\ 
\texttt{RandTask-Poly} & 76.7 & 67.6 \\ 
\texttt{MBC-$\mu$} & 78.8  & 67.0  \\
\texttt{MBC-Poly} & \underline{78.8} & \underline{68.2}\\
\bottomrule
\end{tabular}
\caption{Supervised adaptation results on 12 SNI held-out tasks for Phi-2 obtained in the full (100\% of training data) and limited (10\% of the training data) data settings.}
\label{tab:supervised_adaptation}
\end{table}

\subsection{Summary of Results}
Mirroring the questions at the start of this section, we list our main takeaway messages below:
\begin{enumerate}
\setlength{\itemsep}{-2pt}
\item When appropriately routed, independently trained experts (\texttt{Private-}\arrow{}) can match and surpass the zero-shot performance of full fine-tuning (for Phi-2) and shared tuning (for Mistral 7B). This is a rather surprising result given that experts are independently trained and routing is learned \emph{post-hoc}. These results show promise for building collaboratively and asynchronously trained LMs.
\item If data sharing is possible, then clustering tasks by their similarity with \texttt{MBC} constitutes a very effective strategy. In this case, simply averaging the LoRA adapters obtained through \texttt{MBC} (\texttt{MBC-}$\mu$) is sufficient compared to more sophisticated routing. Our zero-shot and supervised adaptation results underscore the superiority of task-based over example-based clustering. 
\item $\texttt{Arrow}$ appears to be a very performant zero-shot routing strategy while requiring minimal information about the trained LoRAs and none about the training data. For supervised adaptation, training both adapters and the routing coefficients appears to be crucial. Overall, if routing seems beneficial for large libraries of adapters, the gains for smaller libraries are diminishing. This appears to stand in contrast with sparse MoE models, where (non-uniform) routing is crucial~\citep{jiang2024mixtral}. This may be due to the linearity of LoRA experts, which stands in contrast with MLP experts in sparse MoEs~\citep{fedus2021switch}; however, we leave this investigation for future work.
\end{enumerate}

Our main finding is that adapter parameters are suitable both to inform task clustering, and thus guide library building, and to route new inputs, thus facilitating library reuse.

\section{Related Work}
\textbf{Multi-task learning} %
involves training on a joint set of all tasks~\cite{caruana1997multitask}, potentially leading to performance degradation due to task interference~\cite{zhao2018modulation}. An extensive literature studies how to partition learnable parameters into shared and task-specific ones~\cite{ding2023mitigating,strezoski2019many,bragman2019stochastic,zaremoodi2018adaptive,wallingford2022task,fifty2021efficiently}. We operate in the parameter-efficient multi-task learning setting~\citep{ponti2022combining,vu2021spot,chronopoulou2023adaptersoup,pfeiffer2020adapterfusion}.~\citet{vu2021spot} train one prefix adapter~\citep{li-eisner-2019-specializing} per task and learn to re-use them for other tasks based on the adapter similarities. \texttt{MBC} can be seen as an extension of this approach where we cluster tasks based on their weight similarity to ensure more transfer during multi-task pre-training.

\textbf{Mixture of experts} (MoEs), when coupled with sparse routing, are notable for augmenting model capacity with minimal computational overhead \cite{fedus2021switch}. Among the most important differences in this work: i) adapter experts are not trained during base model pre-training, ii) they are parameter-efficient and iii) they are tailored to specific tasks instead of being opaque computation units at the token level whose specialization is not easily interpretable~\citep{jiang2024mixtral}. Regarding ii),~\citet{wang2022adamix,zadouri2023pushing,muqeeth2023soft} employs routing each example to a set of experts, showcasing enhanced performance on unseen tasks. \citet{gupta2022sparsely} trains a separate router for each task and picks a router from a similar task based on domain knowledge. \citet{ye2022eliciting} proposes task-level MoEs that treat a collection of transformer layers as experts and a router chooses from these experts dynamically. Recent work by~\citet{caccia2023multihead,ponti2022combining,ostapenko2023case} investigate the effectiveness of densely routed adapter experts trained end-to-end with an expert library for MTL fine-tuning. For expert aggregation, we employ parameter-space weighted averaging~\cite{wortsman2022model,zhang2023composing,rame2023model} with weights induced by a learned router, a technique akin to those in previous works~\cite{ostapenko2023case,zadouri2023pushing}. Several recent works have also proposed techniques for learning how to route queries to specialized pretrained open-source LLMs~\cite{lu2023routing,shnitzer2023large}.

\textbf{Model ensembling} techniques aim to enhance model robustness and generalization by integrating multiple distinct models~\cite{frankle2020linear,wortsman2022model,rame2023model,jin2022dataless,matena2022merging,chronopoulou2023language,yang2023adamerging}. Parameter space averaging of independent models serves as an efficient ensembling method for full models~\cite{ilharco2022editing,ainsworth2022git,jin2022dataless} and adapters~\cite{zhang2023composing,yadav2024ties}, requiring only a single forward pass through the model, unlike output space ensembling~\cite{dietterich2000ensemble,breiman1996bagging}, that requires many forward passes. Efficient output ensembling techniques that can be applied in conjunction with our work are in~\cite {wen2020batchensemble}. Similarly, \citet{pfeiffer2020adapterfusion} proposes ensembling bottleneck style adapters with the subsequent fine-tuning step. \citet{tam2023merging} presents a merging framework called \texttt{MaTs} using the conjugate gradient method. \citet{yadav2024ties} proposes Ties-Merging to mitigate interference due to redundant parameter values. \citet{daheim2024model} merge models by reducing their individual gradient mismatch with an ideal joint model, weighting their parameters with normalized Fisher Information.

\textbf{Data Clustering for LMs} have been proposed to improve performance and decrease task interference~\cite{fifty2021efficiently,gururangan2023scaling,gou2023mixture}. These methods include clustering using similarities computed by tf-idf and neural embeddings, K-means clustering with balanced linear assignment, and soft clustering with GMMs~\citep{gross2017hard, chronopoulou2023adaptersoup, chronopoulou2021efficient, gururangan2023scaling, duan2021enslm, caron2018deep}. Recent work by \citet{zhou2022efficiently} observes the potential of adapter parameters as effective task embeddings for clustering purposes, a concept we leverage in this work. A similar observation, but regarding task gradients, has been made by~\citet{vu2020exploring}.

\textbf{Building libraries of composable experts} has been envisioned in several previous works \citep{pfeiffer2020adapterfusion,wu2023pi,huang2023lorahub,shah2023ziplora,mole}.~\citet{beck2021adapterhub,poth2023adapters} orchestrated a framework for assembling diverse adapters, offering flexibility in both training and inference. Most related to this work, \citet{huang2023lorahub} build LoRAHub, a library of task-specific LoRAs that can be combined for few-shot generalization. \citet{pfeiffer2020adapterfusion} introduce a two-stage learning algorithm that leverages knowledge from multiple tasks. They first learn task-specific experts and then combine the experts in a separate knowledge composition step. \citet{mole} introduces a learnable gating function to combine multiple LoRAs, called Mixture of LoRA Experts (MoLE). \citet{wu2023pi} presents $\pi$-tuning for vision, language, and vision-language few-shot tasks. $\pi$-tuning trains task-specific experts and then uses task embedding based on the diagonal of the Fisher information matrix to retrieve the top-k most similar tasks to a target task. We extend and complement these works by %
\emph{i)} proposing novel methods to build a library, and \emph{ii)} proposing techniques for zero-shot post-hoc routing independently trained adapters. 
Related to \emph{ii)}, in a concurrent work, ~\citet{muqeeth2024learning} learns a sigmoid gate for each expert, which is later used as expert prototype for zero-shot transfer.
Notably, this method is applicable to the same setting as \texttt{Arrow}, and generalizes beyond linear adapters. However, in contrast to \texttt{Arrow}, obtaining the expert prototypes requires additional training after the experts are learned. %

\section{Conclusions and Future Work}
We investigate how to build and reuse a library of adapters ``end-to-end''. We show the potential of reusing independently (or partially independently) trained adapters with a zero-shot routing strategy. Overall, we strategically investigate the modular augmentation of smaller (language) models, offering a promising direction for research that prioritizes efficiency, flexibility, and performance.

The current investigation focuses on LoRA adapters. For future work, we are excited by the exploration of a heterogeneous ``universe'' of adapters---including soft and hard prompts~\citep{lester2021power,wen2023hard}, MLPs~\citep{houlsby2019parameter}, etc.---and combinations thereof. %
Whether our approach can result in encouraging results at a greater scale (both in terms of data and model size) %
remains open to further investigation. Using the proposed routing strategy for modular continual learning~\cite{ostapenko2021continual,ermis2022memory,wang2022learning} is another promising direction for future work, especially given the fact that the \texttt{Arrow} router is local to each expert. In principle, it may be less susceptible to catastrophic forgetting as no gradient-based training is required to incorporate new experts into the library.

\section{Broader Impact}
This work sheds light on different ways of extending the capabilities of language models by surrounding them with a universe of lightweight adapters that can be trained on conventional hardware. 
Allowing the reuse of adapters might enable systems that are trained in a collaborative and distributed fashion and that use less total energy, with positive ramifications for the environment, but still attain the performance of vanilla systems. Further, this might allow users with smaller computational resources to more easily use and customize LLMs. There are also many potential societal consequences of improving LLMs, some being less desirable and even undesirable, but none of which we feel must be specifically highlighted here.

\section{Summary of Contributions}

\begin{itemize}
\setlength{\itemsep}{-2pt}
    \item OO led the effort on library compression and adaptation baselines. conceptualized and implemented MBC clustering, designed and implemented various experiments, and contributed to paper writing and codebase.
    \item ZS worked on 0-shot and supervised adaptation, designed the task predictor routing, implemented the $\pi$-tuning baseline, and contributed to the codebase and writing.
    \item EP, LCh, NLR were involved in the project after its start, and contributed to the general vision and to proof-writing.
    \item MP maintained and optimized the code, and prepared the code release. 
    \item LCa led the efforts on 0-shot routing; designed, conceptualized, and implemented Arrow routing; implemented the CM baseline and contributed to the codebase.
    \item AS led the project and conceived its idea, worked on the codebase, data generation and evaluations, and wrote the paper.
\end{itemize}

\bibliography{main}
\bibliographystyle{icml2024}

\newpage
\onecolumn
\section{Appendix}

\subsection{Analyzing $\|AB^Tv\|_2$ for in-distribution and out-of-distribution samples}
\label{sec:norm_analysis}

In this section, we analyze whether the motivation behind \texttt{Arrow} routing holds in practice. Recall that at each layer, \texttt{Arrow} routing initializes prototypes in the linear router for expert $i$ with the unit vector $v_i$ maximizing $\|AB^Tv\|_2$. Concretely, we hypothesize that for a hidden activation $h$ computed from $x \in \mathcal{D}_{i}$, we have $\|A_iB_i^Tv\|_2 > \|A_jB_j^Tv\|_2$, for experts $i,j$. In other words, the norm of the linearly transformed prototype will be higher under the expert belonging to the same task as the input $h$. 

To test this hypothesis, we run the following experiment. Let $h_l$ denote the input to the expert at layer $l$, and $(AB^T)_l^i$ denote the linear transformation of expert $i$ at layer $l$. We first sample 5000 examples from the multitask dataset. Then, for a given input $x \in \mathcal{D}_{i}$ at each layer $l$, we compute both $\|(AB^T)_l^i \cdot h_l\|_2 $ and $\|(AB^T)_l^j \cdot h_l\|_2$ where $j$ is another randomly sampled expert such that $i \neq j$. We then compute the average norm ratio $r$ across all layers, i.e. 

$$r = \sum_l^L \frac{1}{L}\frac{\|(AB^T)_l^i \cdot h^i_l\|_2 } {\|(AB^T)_l^j \cdot h^i_l\|_2} .$$

Note that the random expert $j$ is sampled at every layer, and the output of the in-distribution expert is propagated to the next layer. As such, $r > 1$ indicates that on average, the in-distribution expert produces a higher norm output, which would validate the use of the norm-maximizing initialization that \texttt{Arrow} routing uses. In figure \ref{fig:norm_analysis_arrow}, we see that for all the points considered, this ratio is positive, indicating that in-distribution experts tend to be maximize the norm of the linearly transformed input.

\begin{figure}[h]
\centering
\includegraphics[width=0.5\linewidth]{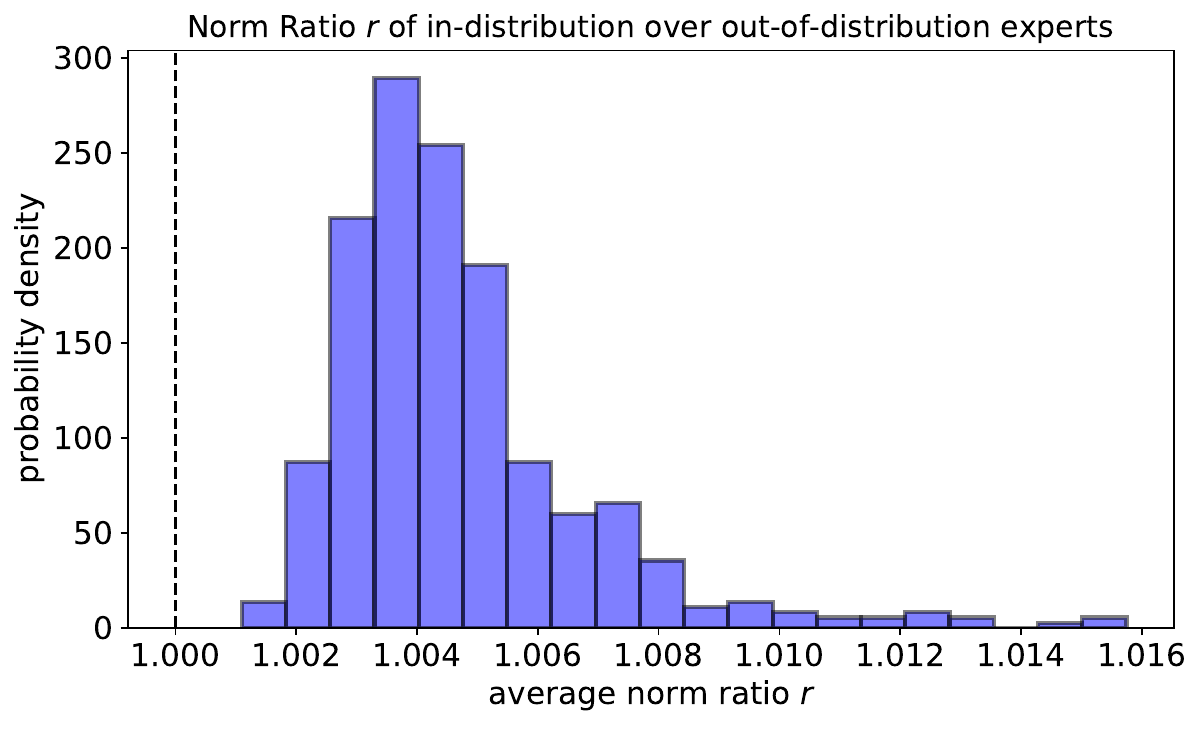}
\caption{Histogram of the ratios $r$ computed over 5000 samples. }
\label{fig:norm_analysis_arrow}
\end{figure}

\begin{table*}[t!]
\scriptsize
\centering
\resizebox{0.95\textwidth}{!}{
\begin{tabular}{llcc|cccccccccc|c} 
\toprule
&\texttt{Library} &  & L&\texttt{piqa} & \texttt{boolq} & \texttt{wgrande} & \texttt{hswag} & \texttt{arcE} & \texttt{arcC} & \texttt{HE} & \texttt{oqa} & \texttt{bbh} & \texttt{mbpp} & $\texttt{Acc}.$ \\
\midrule
\midrule
\multirow{6}{*}{\rotatebox[origin=c]{90}{\emph{\textbf{StableLM (3B) }}}} & \texttt{Base} & - & - &78.2 & 73.1 & 66.6 & 73.7 & 59.6 & 41.5 & \dashuline{18.3} &  37.6 & 34.7 & 32.3 & 51.6 \\ 
&\texttt{Shared} & - & 1& 79.4 & 80.3 & 68.0 & 71.3 & 74.7 & 42.1 & 11.6 & 38.0 & 38.3 & 21.0 & 52.5 \\
\cmidrule(lr){2-15}
& \texttt{Private} & $\mu$ &  100 & 79.4 & 76.8 & 67.3 & 74.4 & 72.4 & 44.0 & 16.5 & 42.6 & 37.0 & 34.6 & 54.5\\
& \texttt{Private} & \arrow{} & 100& 80.1 & 72.1 & 70.8 & 74.8 & 73.4 & 45.3 & 16.5 & 43.6 & 36.1 & 33.5 & 54.6\\
\cmidrule(lr){2-15}
& \texttt{MBC} & $\mu$ &10& 80.4 & 80.4 & 68.2 & 74.7 & 76.7 & 47.4 & 14.6 & 43.0 & 35.4 & 36.2 & \underline{55.7} \\
& \texttt{MBC} & $\arrow{}$ &10& 80.5 & 79.0 & 68.2 & 73.6 & 75.2 & 46.4 & 13.4 & 43.0 & 32.0 & 27.6 & 53.9 \\
\bottomrule
\end{tabular}
}
\caption{\label{tab:zeroshot_stablelm}\textbf{Out-of-distribution zero-shot results}: Accuracy on held-out tasks for StableLM. The best results are underlined.
}
\end{table*}

\begin{table*}[htbp]
\centering
\resizebox{0.95\textwidth}{!}{
  \begin{small}
\begin{tabular}{llc|cccccccccccc|c}
\toprule
&\texttt{Method} & $L$ & \multicolumn{12}{c|}{\texttt{SNI Tasks}} & \texttt{Rouge-L} \\
& & & \texttt{202} & \texttt{304} & \texttt{614} & \texttt{613} & \texttt{362} & \texttt{242} & \texttt{1728} & \texttt{1557} & \texttt{035} & \texttt{1356} & \texttt{039} & \texttt{1153} & \\
\midrule
\multirow{7}{*}{\rotatebox[origin=c]{90}{\emph{\textbf{Phi-2 (2.8B) }}}} & \texttt{Base} & -&4.0&3.3&26.4&3.5&16.2&32.5&35.2&62.5&54.2&12.8&8.2&7.6&22.2 \\
\cmidrule(lr){2-16}
&\texttt{Shared} & 1 & 38.3&17.9&36.4&11.5&77.2&39.4&45.8&84.5&40.7&21.5&34.3&24.1&39.3\\
\cmidrule(lr){2-16}
&\texttt{Private}-$\mu$ & 256 & 10.6&16.1&35.6&9.6&64.8&58.2&42.6&72.2&61.7&17.5&25.1&24.0& 36.6 \\
&\texttt{Private-$\arrow{}$} & 256 & 20.4 & 18.8 & 31.8 & 10.5 & 76.3 & 36.4 & 46.8 & 84.2 & 41.8 & 19.2 & 33.4 & 28.7 & 37.4 \\
\cmidrule(lr){2-16}
&\texttt{MBC}-$\mu$ & 10 & 31.8&26.9&33.9&12.7&77.6&77.9&47.2&86.0&49.2&22.4&37.0&29.8&\underline{44.4} \\   
&\texttt{MBC-$\arrow{}$} & 10 & 32.6 & 15.9 & 31.3 & 7.6 & 79.6 & 36.6 & 41.7 & 80.2 & 33.1 & 21.5 & 32.0 & 28.5 & 36.7 \\ 
\midrule
\midrule
\multirow{8}{*}{\rotatebox[origin=c]{90}{\emph{\textbf{Mistral (7B) }}}} & \texttt{Base} & -& 13.7 & 10.7 & 31.8 & 5.6 & 37.4 & 22.0 & 35.6 & 49.1 & 58.6 & 13.7 & 22.2 & 14.2 & 26.4 \\
\cmidrule(lr){2-16}
&\texttt{Shared} & 1 & 50.8 & 18.0 & 37.5 & 8.8 & 67.4 & 80.0 & 54.1 & 81.9 & 59.6 & 30.0 & 32.0 & 27.0 & 45.6 \\
\cmidrule(lr){2-16}
&\texttt{Private}-$\mu$ & 256 & 30.1 & 17.2 & 10.2 & 7.7 & 70.4 & 37.7 & 38.6 & 63.0 & 63.0 & 20.7 & 25.4 & 23.2 & 36.4 \\
&\texttt{Private-$\arrow{}$} & 256 & 38.5 & 23.7 & 43.7 & 12.7 & 78.0 & 76.7 & 54.6 & 83.3 & 57.9 & 25.2 & 35.4 & 33.4 & 46.9\\
\cmidrule(lr){2-16}
&\texttt{MBC}-$\mu$ & 10 & 54.0 & 26.1 & 46.4 & 15.0 & 80.8 & 80.1 & 46.0 & 82.7 & 66.5 & 28.1 & 46.1 & 36.0 & \underline{50.6} \\   
&\texttt{MBC-$\arrow{}$} & 10 & 38.1 & 24.3 & 35.9 & 12.8 & 85.5 & 77.1 & 43.3 & 82.1 & 57.7 & 29.0 & 33.9 & 31.2 & 45.9 \\ 
\bottomrule
\end{tabular}
\end{small} 
}
\caption{\label{ref:sni-zero-shot}\textbf{Out-of-distribution zero-shot results on 12 held-out SNI tasks} for library built for the Phi-2 and Mistral base models. Applying $\arrow{}$ routing to the Private libraries results in performance improvements over the $\mu$ routing for both models, with a notable improvement of over 10 Rouge-L points in case of Mistral. It is worth noticing that $\mu$ routing performed better than $\arrow$ in case of MBC library for both models. We note that $\arrow$ only selects top-4 experts for routing, whereas $\mu$ averages full libraries. Best results are underlined.}
  \label{tab:zero_shot_sni_phi2}
\end{table*}

\subsection{Few-shot adaptation}
\label{sec:data_scarse}
We apply some of the proposed methods to a data scarce setting with up to only 0.5\% of the original training data per task (approx. 40 examples per task). We show the results in Table~\ref{tab:scarce_Data}. Even in this setting gradient based method \texttt{MBC-Poly} considerably outperforms \texttt{LoraHub}, where the \texttt{LoraHub} is given compute equivalent to training gradient based methods on full dataset. Additionally, we observe that \texttt{MBC-PolyZ}, a method similar to \texttt{MBC-Poly} that only updates the routings and not the expert's weights, performs similarly to \texttt{LoraHub}. Interestingly, when data amount is lowered, the perfromance of \texttt{MBC-PolyZ} is reduced by a relatively smaller margin than \texttt{MBC-Poly} which can be explained by a smaller amount of updated parameters.

\begin{table*}[htbp]
\centering
\resizebox{0.95\textwidth}{!}{
\begin{small}
\begin{tabular}{lc|cccccccccccc|c}
\toprule
\texttt{Method} & $L$ & \multicolumn{12}{c|}{\texttt{SNI Tasks}} & \texttt{Rouge-L} \\
& & \texttt{202} & \texttt{304} & \texttt{614} & \texttt{613} & \texttt{362} & \texttt{242} & \texttt{1728} & \texttt{1557} & \texttt{035} & \texttt{1356} & \texttt{039} & \texttt{1153} & \\
\midrule
\midrule
\midrule
\emph{Full data}\\
\texttt{MBC-LoraHub} & 10 & 41.5 & 21.9 & 37.4 & 17.5 & 78.1 & 68.3 & 48.0 & 82.0 & 62.6 & 21.2 & 33.5 & 31.1 & 45.3 \\
\texttt{MBC-Poly} & 10 & 96.9 & 84.4 & 67.2 & 53.9 & 96.4 & 97.8 & 60.2 & 87.9 & 91.3 & 29.4 & 81.7 & 99.7 & 78.9 \\
\texttt{MBC-PolyZ} & 10 & 37.7 & 27.9 & 36.2 & 12.6 & 75.9 & 74.4 & 48.7 & 81.3 & 58.9 & 22.5 & 36.1 & 31.1 & 45.3 \\

\midrule
\emph{10\%} \\
\texttt{MBC-Poly} & 10 & 89.6 & 53.3 & 64.5 & 44.5 & 93.5 & 98.5 & 58.5 & 75.7 & 87.3 & 27.2 & 65.6 & 66.8 & 68.8 \\
\texttt{MBC-PolyZ} & 10 & 34.3 & 27.5 & 36.2 & 12.4 & 76.5 & 74.3 & 47.5 & 86.2 & 57.9 & 22.7 & 35.3 & 31.4 & 45.2 \\

\midrule
\emph{5\%} \\
\texttt{MBC-Poly} & 10 & 87.0 & 43.0 & 61.3 & 41.7 & 92.0 & 95.2 & 55.3 & 77.3 & 89.0 & 25.4 & 59.1 & 47.8 & 64.5 \\
\texttt{MBC-PolyZ} & 10 & 32.6 & 28.6 & 36.0 & 13.0 & 76.4 & 73.9 & 47.3 & 86.3 & 57.7 & 22.7 & 36.6 & 31.0 & 45.2  \\
\midrule
\emph{0.5\%} \\
\texttt{MBC-Poly} & 10 & 49.7 & 30.4 & 43.7 & 20.3 & 77.3 & 78.4 & 48.2 & 86.5 & 72.2 & 23.2 & 43.0 & 29.1 & 50.2  \\
\texttt{MBC-PolyZ} & 10 & 32.0 & 27.4 & 34.3 & 12.6 & 77.6 & 78.1 & 47.2 & 86.5 & 53.2 & 22.2 & 37.0 & 29.1 & 44.8  \\
\bottomrule
\end{tabular}
\end{small}  
}
\caption{\label{tab:supervised2}Rouge-L score for Phi-2 model after adaptation with different portions of data per task ranging from full dataset down to 10\% and 5\% of data per task. Note, \texttt{MBC-PolyZ} only tunes the routing weights, whereas \texttt{MBC-Poly} trains both the routing weights and the experts parameters.}
  \label{tab:scarce_Data}
\end{table*}

\begin{table*}[htbp]
\centering
\resizebox{0.95\textwidth}{!}{
  \begin{small}
\begin{tabular}{llc|cccccccccccc|c}
\toprule
&\texttt{Method} & $L$ & \multicolumn{12}{c|}{\texttt{SNI Tasks}} & \texttt{Rouge-L} \\
& & & \texttt{202} & \texttt{304} & \texttt{614} & \texttt{613} & \texttt{362} & \texttt{242} & \texttt{1728} & \texttt{1557} & \texttt{035} & \texttt{1356} & \texttt{039} & \texttt{1153} & \\
\midrule
\midrule
\multirow{8}{*}{\rotatebox[origin=c]{90}{\emph{\textbf{Phi-2 (2.8B) }}}}&\texttt{No Library} 
& - & 93.2 & 74.2 & 64.9 & 51.4 & 95.9 & 96.2 & 59.3 & 81.4 & 90.5 & 26.9 & 73 & 99.1 & 75.5 \\
\cmidrule(lr){2-16}
&\texttt{Shared} & 1 &  93.1 & 73.4 & 65.0 & 48.9 & 96.0 & 95.9 & 58.4 & 86.8 & 91.2 & 29.0 & 73.4 & 98.4 & 75.8 \\
&\texttt{MHR} & 1 & 94.5 & 66.9 & 63.0 & 47.8 & 94.9 & 95.6 & 59.5 & 86.6 & 91.0 & 27.9 & 70.7 & 98.2 & 74.8 \\
&\texttt{Poly} & 10 & 92.1 & 66.4 & 63.0 & 45.9 & 94.9 & 96.4 & 56.0 & 85.3 & 90.2 & 27.6 & 70.0 & 93.2 & 73.4 \\
\cmidrule(lr){2-16}
&\texttt{Private}-$\mu$ & 256 & 93.6 & 78.1 & 65.0 & 50.7 & 94.8 & 97.8 & 59.7 & 87.9 & 90.7 & 28.1 & 76.4 & 99.7 & 76.9 \\
&\texttt{MBC}-$\mu$ & 10 & 96.4 & 83.2 & 67.6 & 53.5 & 96.2 & 98.0 & 60.5 & 88.2 & 90.7 & 29.8 & 82.3 & 99.5 & 78.8 \\
\cmidrule(lr){2-16}
&\texttt{MBC-LoraHub} & 10 & 41.5 & 21.9 & 37.4 & 17.5 & 78.1 & 68.3 & 48.0 & 82.0 & 62.6 & 21.2 & 33.5 & 31.1 & 45.3 \\
\cmidrule(lr){2-16}
&\texttt{RandTask-Poly} & 10 & 96.4 & 77.1 & 66.5 & 48.6 & 96.7 & 98.9 & 59.9 & 85.1 & 90.7 & 28.6 & 73.9 & 97.5 & 76.7  \\
&\texttt{MBC-Poly} & 10 & 96.9 & 84.4 & 67.2 & 53.9 & 96.4 & 97.8 & 60.2 & 87.9 & 91.3 & 29.4 & 81.7 & 99.7 & \underline{78.9} \\
\midrule
\midrule
\multirow{8}{*}{\rotatebox[origin=c]{90}{\emph{\textbf{Mistral (7B)}}}}&\texttt{No Library} & - & 97.6 & 88.3 & 68.9 & 59.9 & 98.8 & 98.8 & 62.9 & 87.3 & 91.8 & 37.5 & 80.5 & 100 & 81.0 \\
&\texttt{Shared} & 1 & 95.8 & 87.4 & 69.9 & 52.7 & 98.7 & 99.2 & 63.5 & 87.6 & 91.6 & 37.2 & 78.1 & 100 & 80.1 \\
\cmidrule(lr){2-16}
&\texttt{Private-$\mu$} & 256 & 98.5 & 87.2 & 70.6 & 54.1 & 98.3 & 99.1 & 64.0 & 89.1 & 92.0 & 37.3 & 81.0 & 100 & \underline{80.9} \\ 
&\texttt{MBC-$\mu$} & 10 & 98.1 & 84.8 & 70.1 & 54.2 & 98.7 & 95.7 & 62.8 & 82.9 & 92.0 & 38.2 & 82.1 &99.5 & 79.9 \\
\cmidrule(lr){2-16}
&\texttt{MBC-LoRAHub} & 10 & 47.8 & 23.1 & 45.9 & 14.0 & 81.4 & 79.6 & 49.7 & 84.9 & 69.6 & 28.6 & 42.2 & 34.5 & 50.1 \\ 

\cmidrule(lr){2-16}
&\texttt{RandTask-Poly} & 10 & 98.4 & 86.9 & 69.3 & 53.8 & 96.1 & 93.0 & 64.7 & 84.5 & 92.1 & 39.3 & 80.1 & 99.5 & 79.8\\ 
&\texttt{MBC-Poly} & 10 & 98.7 & 88.7 & 69.2 & 56.1 & 97.4 & 99.5 & 64.1 & 82.2 & 92.2 & 38.7 & 81.1 & 99.0 & 80.6 \\
    
\bottomrule
\end{tabular}
\end{small} 
}
\caption{\textbf{Supervised  adaptation results (100\% training data per task)}: Rouge-L on 12 held-out SNI for Phi-2 and Mistral 7B models for different libraries. \texttt{LoraHub} follows the original implementation and optimizes the weighting coefficients for the adapters in the library with a non-gradient based optimizer. Best results are underlined.}
  \label{tab:adaptation_results_100}
\end{table*}

\begin{table*}[htbp]
\centering
\resizebox{0.95\textwidth}{!}{
  \begin{small}
\begin{tabular}{llc|cccccccccccc|c}
\toprule
&\texttt{Method} & $L$ & \multicolumn{12}{c|}{\texttt{SNI Tasks (10\%)}} & \texttt{Rouge-L} \\
& & & \texttt{202} & \texttt{304} & \texttt{614} & \texttt{613} & \texttt{362} & \texttt{242} & \texttt{1728} & \texttt{1557} & \texttt{035} & \texttt{1356} & \texttt{039} & \texttt{1153} & \\
\midrule
\midrule
\multirow{8}{*}{\rotatebox[origin=c]{90}{\emph{\textbf{Phi-2 (2.8B) }}}}&\texttt{No Library} & - & 71.5 & 36.1 & 53.6 & 36.9 & 80.0 & 85.5 & 46.7 & 62.3 & 84.0 & 21.7 & 41.3 & 27.2 & 53.9 \\
\cmidrule(lr){2-16}
&\texttt{Shared} & 1 & 76.8 & 35.5 & 55.5 & 39.4 & 82.6 &  89.3 & 47.6 & 61.8 & 86.0 & 23.3 & 47.9 & 31.2 & 56.4 \\
&\texttt{MHR} & 1 & 83.6 & 45.1 & 58.2 & 40.0 & 91.3 & 94.3 & 54.0 & 84.1 & 85.7 & 25.7 & 58.0 & 54.2 & 64.5 \\
&\texttt{Poly} & 1 & 74.4 & 38.3 & 57.8 & 39.5 & 82.5 & 92.2 & 50.8 & 85.1 & 85.1 & 25.5 & 54.8 & 54.1 & 61.7 \\
&\texttt{Private}-$\mu$ & 256 & 81.7 & 41.2 & 60.6 & 40.4 & 89.8 & 96.0 & 49.6 & 75.1 & 87.1 & 23.8 & 57.2 & 47.9 & 62.5 \\
&\texttt{MBC}-$\mu$ & 10 & 86.2 & 52.3 & 64.3 & 43.8 & 93.8 & 97.3 & 53.3 & 75.0 & 87.5 & 26.3 & 61.1 & 63.8 & 67.0 \\
\cmidrule(lr){2-16}
&\texttt{MBC-LoraHub} & 10 & 43.6 & 22.2 & 36.5 & 13.5 & 77.0 & 68.8 & 45.5 & 82.2 & 63.2 & 21.2 & 34.6 & 27.6 & 44.7 \\
\cmidrule(lr){2-16}
&\texttt{RandTask-Poly} & 10 & 87.9 & 51.0 & 63.5 & 41.4 & 94.1 & 95.8 & 55.6 & 79.6 & 89.0 & 27.1 & 61.1 & 65.3 & 67.6\\
&\texttt{MBC-Poly} & 10 & 88.9 & 52.0 & 64.4 & 45.6 & 94.3 & 96.9 & 56.7 & 75.2 & 87.5 & 27.1 & 64.4 & 66.0 & \underline{68.2} \\
\midrule
\midrule
\multirow{8}{*}{\rotatebox[origin=c]{90}{\emph{\textbf{Mistral (7B)}}}}&\texttt{No Library} & - & 91.7 & 66.8 & 66.2 & 47.8 & 95.2 & 98.3 & 59.5 & 69.9 & 90.7 & 33.9 & 65.8 & 68.1 & 71.2 \\
\cmidrule(lr){2-16}
&\texttt{Shared} & 1 & 94.6 & 64.9 & 65.1 & 45.3 & 90.7 & 91.0 & 60.3 & 82.7 & 89.6 & 33.4 & 66.3 & 91.3 & 72.9  \\
\cmidrule(lr){2-16}
&\texttt{Private-$\mu$} & 256 & 89.0 & 64.3 & 66.0 & 47.2 & 94.7 & 98.8 & 59.4 & 84.1 & 90.6 & 33.5 & 67.6 & 92.9 & 74.0 \\ 
&\texttt{MBC-$\mu$} & 10 & 94.2 & 62.6 & 66.3 & 47.9 & 95.9 & 96.6 & 59.9 & 83.6 & 90.7 & 33.7 & 69.3 & 92.8 &  74.5 \\
\cmidrule(lr){2-16}
&\texttt{MBC-LoRAHub} & 10 & 47.7 & 23.2 & 47.0 & 15.4 & 85.8 & 72.4 & 49.0 & 81.9 & 71.7 & 27.3 & 27.7 & 30.6 & 48.3 \\ 
\cmidrule(lr){2-16}
&\texttt{RandTask-Poly} & 10 & 95.0 & 64.8 & 66.0 & 48.8 & 94.3 & 92.1 & 59.6 & 87.1 & 90.6 & 34.4 & 66.4 & 92.4 & 74.3 \\ 
&\texttt{MBC-Poly} & 10 & 95.1 & 66.3 & 66.0 & 48.3 & 96.4 & 98.4 & 60.2 & 84.9 & 90.5 & 33.7 & 67.8 & 92.7 & \underline{75.0} \\
    
\bottomrule
\end{tabular}
\end{small} 
}
\caption{\textbf{Supervised few-shot adaptation results (10\% training data per task)}: Rouge-L on 12 held-out SNI for Phi-2 and Mistral 7B models for different libraries. \texttt{LoraHub} follows the original implementation and optimizes the weighting coefficients for the adapters in the library with a non-gradient based optimizer. Best results are underlined.}
  \label{tab:adaptation_results_10}
\end{table*}

\section{Implementation details and hyperparameters}
In this section we provide some technical details about the experiments conducted in this paper.

\textbf{Training hyperparameters.} For all LoRA experts trained in this paper we employ LoRA rank of 4, LoRA dropout probability of 0.05, LoRA $\alpha$ of 16, and a learning rate of 1e-4 with a learning rate warm-up and annealing phases.  We experimented with only patching fully connected layers (FC), only attention layers + attention output projection (ATT+O) or both (BOTH). For the preliminary experiments in Figure~\ref{fig:interf} we modify only the MLP (FC) layers of the transformer (.*fc[12].*). We found that patching FC layers severely underperform ATT+O layers. Patching BOTH gives marginal gains over ATT+O while significantly increasing computation cost and memory usage due to the wide projection (4 * hidden size) of the first FC layer in the transformer residual block. Therefore, for the rest of the experiments, we modified attention layers + attention output projection (e.g. .*Wqkv.* $|$.*out\_proj.* for Phi-2).

\textbf{Downstream zero-shot results.} All library-bases downstream zero-shot results are reported using top-4 routing with temperature 1. Unless stated otherwise, for all \texttt{MBC} libraries we use 10 experts. Additionally, in our implementation of the downstream evaluation we append an EOS token to the target options to mark the end of a sentence. We use token-length normalized scores for selecting continuations for multiple-choice tasks evaluation~\cite{eleutherai2021multiple}.

\textbf{Adaptation experiments.} For the adaptation experiments we also use the learning rate of 1e-4, with the same learning rate schedule as stated above. For both \texttt{MHR} and \texttt{Poly} adaptation, we tune both the experts and the routing weights.

\begin{table*}[htbp]
\centering
\scriptsize
\begin{tabularx}{\textwidth}{|l|X}
\toprule
c0 & 
        "ropes\_background\_new\_situation\_answer",
        "ropes\_prompt\_bottom\_no\_hint",
        "ropes\_plain\_background\_situation",
        "ropes\_new\_situation\_background\_answer",
        "ropes\_given\_background\_situation",
        "ropes\_prompt\_bottom\_hint\_beginning",
        "ropes\_prompt\_beginning",
        "ropes\_read\_background\_situation",
        "ropes\_plain\_bottom\_hint",
        "ropes\_plain\_no\_background",
        "ropes\_prompt\_mix",
        "ropes\_background\_situation\_middle" \\
\midrule    
c1 & 
        "glue\_sst2\_2\_0\_0",
        "adversarial\_qa\_droberta\_generate\_question",
        "true\_case",
        "stream\_qed",
        "huggingface\_xsum",
        "cot\_esnli",
        "cot\_gsm8k",
        "trec\_1\_0\_0",
        "yelp\_polarity\_reviews\_0\_2\_0",
        "lambada\_1\_0\_0",
        "glue\_cola\_2\_0\_0",
        "ag\_news\_subset\_1\_0\_0",
        "gem\_dart\_1\_1\_0",
        "math\_dataset\_algebra\_\_linear\_1d\_1\_0\_0",
        "cnn\_dailymail\_3\_4\_0",
        "wiki\_hop\_original\_explain\_relation",
        "dbpedia\_14\_given\_list\_what\_category\_does\_the\_paragraph\_belong\_to",
        "gem\_wiki\_lingua\_english\_en\_1\_1\_0",
        "fix\_punct",
        "imdb\_reviews\_plain\_text\_1\_0\_0",
        "race\_middle\_Write\_a\_multi\_choice\_question\_for\_the\_following\_article",
        "gigaword\_1\_2\_0",
        "dbpedia\_14\_given\_a\_list\_of\_category\_what\_does\_the\_title\_belong\_to",
        "gem\_web\_nlg\_en\_1\_1\_0",
        "word\_segment",
        "race\_high\_Write\_a\_multi\_choice\_question\_for\_the\_following\_article",
        "wmt16\_translate\_de\_en\_1\_0\_0",
        "cot\_ecqa",
        "aeslc\_1\_0\_0",
        "dream\_generate\_first\_utterance",
        "wmt16\_translate\_fi\_en\_1\_0\_0",
        "dream\_answer\_to\_dialogue",
        "para\_crawl\_enes",
        "adversarial\_qa\_dbert\_generate\_question",
        "race\_middle\_Write\_a\_multi\_choice\_question\_options\_given\_",
        "wmt14\_translate\_fr\_en\_1\_0\_0"\\
\midrule    
c2 & 
        "adversarial\_qa\_dbidaf\_question\_context\_answer",
        "super\_glue\_record\_1\_0\_2",
        "wiki\_hop\_original\_generate\_object",
        "adversarial\_qa\_droberta\_tell\_what\_it\_is",
        "dbpedia\_14\_given\_a\_choice\_of\_categories\_",
        "wiki\_hop\_original\_choose\_best\_object\_affirmative\_3",
        "quac\_1\_0\_0",
        "wiki\_hop\_original\_choose\_best\_object\_interrogative\_1",
        "wiki\_hop\_original\_choose\_best\_object\_affirmative\_1",
        "adversarial\_qa\_dbert\_answer\_the\_following\_q",
        "wiki\_hop\_original\_choose\_best\_object\_interrogative\_2",
        "adversarial\_qa\_droberta\_question\_context\_answer",
        "squad\_v2\_0\_3\_0\_0",
        "wiki\_hop\_original\_generate\_subject",
        "wiki\_bio\_guess\_person",
        "adversarial\_qa\_dbidaf\_answer\_the\_following\_q",
        "adversarial\_qa\_droberta\_answer\_the\_following\_q",
        "adversarial\_qa\_dbert\_tell\_what\_it\_is",
        "race\_high\_Write\_a\_multi\_choice\_question\_options\_given\_",
        "wiki\_hop\_original\_choose\_best\_object\_affirmative\_2",
        "wiki\_hop\_original\_generate\_subject\_and\_object",
        "drop\_2\_0\_0",
        "adversarial\_qa\_dbert\_question\_context\_answer",
        "adversarial\_qa\_dbidaf\_tell\_what\_it\_is"\\
\midrule    
c3 & 
        "wiqa\_what\_might\_be\_the\_first\_step\_of\_the\_process",
        "wiqa\_what\_is\_the\_final\_step\_of\_the\_following\_process",
        "wmt16\_translate\_ro\_en\_1\_0\_0",
        "wiqa\_what\_might\_be\_the\_last\_step\_of\_the\_process",
        "wiki\_bio\_key\_content",
        "gem\_common\_gen\_1\_1\_0",
        "duorc\_SelfRC\_build\_story\_around\_qa",
        "app\_reviews\_generate\_review",
        "wiki\_bio\_what\_content",
        "wiki\_bio\_who",
        "gem\_e2e\_nlg\_1\_1\_0",
        "cot\_esnli\_ii",
        "wmt16\_translate\_tr\_en\_1\_0\_0",
        "wiqa\_what\_is\_the\_missing\_first\_step",
        "wiki\_bio\_comprehension",
        "coqa\_1\_0\_0",
        "duorc\_ParaphraseRC\_build\_story\_around\_qa",
        "multi\_news\_1\_0\_0"\\
\midrule    
c4 & 
        "wiki\_qa\_found\_on\_google",
        "app\_reviews\_categorize\_rating\_using\_review",
        "race\_middle\_Is\_this\_the\_right\_answer",
        "super\_glue\_cb\_1\_0\_2",
        "wiki\_qa\_Topic\_Prediction\_Answer\_Only",
        "wiki\_qa\_Direct\_Answer\_to\_Question",
        "super\_glue\_wsc\_fixed\_1\_0\_2",
        "cot\_gsm8k\_ii",
        "unified\_qa\_science\_inst",
        "race\_high\_Is\_this\_the\_right\_answer",
        "cot\_strategyqa",
        "cot\_ecqa\_ii",
        "quarel\_do\_not\_use",
        "wiki\_qa\_exercise",
        "wiki\_qa\_automatic\_system",
        "cot\_creak\_ii",
        "quarel\_heres\_a\_story",
        "quarel\_choose\_between",
        "stream\_qed\_ii",
        "wiki\_qa\_Topic\_Prediction\_Question\_Only",
        "glue\_qnli\_2\_0\_0",
        "cot\_sensemaking\_ii",
        "super\_glue\_copa\_1\_0\_2",
        "social\_i\_qa\_Generate\_the\_question\_from\_the\_answer",
        "social\_i\_qa\_Show\_choices\_and\_generate\_index",
        "quarel\_testing\_students",
        "wiki\_qa\_Topic\_Prediction\_Question\_and\_Answer\_Pair",
        "wiki\_qa\_Decide\_good\_answer",
        "wiki\_qa\_Jeopardy\_style",
        "wiki\_qa\_Generate\_Question\_from\_Topic",
        "definite\_pronoun\_resolution\_1\_1\_0",
        "wiqa\_effect\_with\_label\_answer",
        "glue\_wnli\_2\_0\_0",
        "cot\_qasc",
        "cot\_strategyqa\_ii",
        "quarel\_logic\_test",
        "stream\_aqua\_ii"\\
\midrule    
c5 & 
        "quoref\_Context\_Contains\_Answer",
        "duorc\_SelfRC\_generate\_question\_by\_answer",
        "quoref\_Find\_Answer",
        "duorc\_ParaphraseRC\_movie\_director",
        "duorc\_ParaphraseRC\_answer\_question",
        "quoref\_Found\_Context\_Online",
        "quoref\_Read\_And\_Extract\_",
        "duorc\_ParaphraseRC\_title\_generation",
        "duorc\_ParaphraseRC\_decide\_worth\_it",
        "quoref\_What\_Is\_The\_Answer",
        "duorc\_ParaphraseRC\_generate\_question",
        "quoref\_Guess\_Title\_For\_Context",
        "quoref\_Answer\_Test",
        "duorc\_SelfRC\_question\_answering",
        "duorc\_SelfRC\_title\_generation",
        "duorc\_ParaphraseRC\_generate\_question\_by\_answer",
        "duorc\_ParaphraseRC\_extract\_answer",
        "duorc\_SelfRC\_answer\_question",
        "duorc\_SelfRC\_decide\_worth\_it",
        "duorc\_ParaphraseRC\_question\_answering",
        "quoref\_Answer\_Question\_Given\_Context",
        "duorc\_SelfRC\_extract\_answer",
        "quoref\_Guess\_Answer",
        "quoref\_Answer\_Friend\_Question",
        "duorc\_SelfRC\_movie\_director",
        "duorc\_SelfRC\_generate\_question",
        "quoref\_Given\_Context\_Answer\_Question"\\
\midrule    
c6 & 
        "super\_glue\_rte\_1\_0\_2",
        "cot\_sensemaking",
        "super\_glue\_wic\_1\_0\_2",
        "cos\_e\_v1\_11\_rationale",
        "anli\_r3\_0\_1\_0",
        "dream\_generate\_last\_utterance",
        "paws\_wiki\_1\_1\_0",
        "cos\_e\_v1\_11\_generate\_explanation\_given\_text",
        "cot\_creak",
        "stream\_aqua",
        "snli\_1\_1\_0",
        "cos\_e\_v1\_11\_i\_think",
        "glue\_qqp\_2\_0\_0",
        "cos\_e\_v1\_11\_explain\_why\_human",
        "anli\_r2\_0\_1\_0",
        "anli\_r1\_0\_1\_0",
        "glue\_stsb\_2\_0\_0",
        "cos\_e\_v1\_11\_aligned\_with\_common\_sense",
        "glue\_mnli\_2\_0\_0",
        "social\_i\_qa\_I\_was\_wondering",
        "cosmos\_qa\_1\_0\_0",
        "glue\_mrpc\_2\_0\_0",
        "social\_i\_qa\_Generate\_answer"\\
\midrule    
c7 & 
        "dream\_read\_the\_following\_conversation\_and\_answer\_the\_question",
        "app\_reviews\_convert\_to\_star\_rating",
        "cos\_e\_v1\_11\_question\_option\_description\_text",
        "social\_i\_qa\_Show\_choices\_and\_generate\_answer",
        "quartz\_answer\_question\_based\_on",
        "sciq\_Direct\_Question\_Closed\_Book\_",
        "qasc\_qa\_with\_separated\_facts\_3",
        "quartz\_given\_the\_fact\_answer\_the\_q",
        "quartz\_answer\_question\_below",
        "kilt\_tasks\_hotpotqa\_final\_exam",
        "sciq\_Multiple\_Choice",
        "wiqa\_does\_the\_supposed\_perturbation\_have\_an\_effect",
        "cos\_e\_v1\_11\_question\_description\_option\_text",
        "wiki\_qa\_Is\_This\_True\_",
        "quartz\_use\_info\_from\_question\_paragraph",
        "sciq\_Direct\_Question",
        "qasc\_qa\_with\_separated\_facts\_2",
        "wiqa\_which\_of\_the\_following\_is\_the\_supposed\_perturbation",
        "app\_reviews\_convert\_to\_rating",
        "cos\_e\_v1\_11\_question\_option\_description\_id",
        "wiqa\_effect\_with\_string\_answer",
        "qasc\_qa\_with\_separated\_facts\_5",
        "dream\_baseline",
        "quartz\_having\_read\_above\_passage",
        "cos\_e\_v1\_11\_question\_description\_option\_id",
        "qasc\_qa\_with\_separated\_facts\_1",
        "cos\_e\_v1\_11\_description\_question\_option\_text",
        "qasc\_qa\_with\_combined\_facts\_1",
        "qasc\_is\_correct\_1",
        "cos\_e\_v1\_11\_description\_question\_option\_id",
        "social\_i\_qa\_Check\_if\_a\_random\_answer\_is\_valid\_or\_not",
        "sciq\_Multiple\_Choice\_Closed\_Book\_",
        "quartz\_use\_info\_from\_paragraph\_question",
        "qasc\_is\_correct\_2",
        "qasc\_qa\_with\_separated\_facts\_4",
        "quartz\_read\_passage\_below\_choose",
        "quartz\_paragraph\_question\_plain\_concat",
        "sciq\_Multiple\_Choice\_Question\_First"\\
\midrule    
c8 & 
        "race\_middle\_Read\_the\_article\_and\_answer\_the\_question\_no\_option\_",
        "race\_high\_Select\_the\_best\_answer",
        "quail\_description\_context\_question\_answer\_id",
        "quail\_context\_question\_description\_text",
        "race\_high\_Read\_the\_article\_and\_answer\_the\_question\_no\_option\_",
        "race\_high\_Select\_the\_best\_answer\_no\_instructions\_",
        "quail\_context\_description\_question\_answer\_id",
        "race\_high\_Taking\_a\_test",
        "super\_glue\_multirc\_1\_0\_2",
        "race\_middle\_Select\_the\_best\_answer",
        "quail\_context\_question\_description\_answer\_id",
        "quail\_description\_context\_question\_answer\_text",
        "quail\_context\_question\_answer\_description\_text",
        "race\_high\_Select\_the\_best\_answer\_generate\_span\_",
        "race\_middle\_Select\_the\_best\_answer\_generate\_span\_",
        "quail\_context\_question\_answer\_description\_id",
        "quail\_context\_description\_question\_answer\_text",
        "quail\_context\_description\_question\_text",
        "quail\_context\_question\_description\_answer\_text",
        "quail\_description\_context\_question\_text",
        "race\_middle\_Taking\_a\_test",
        "quail\_no\_prompt\_id",
        "quail\_no\_prompt\_text",
        \newline"race\_middle\_Select\_the\_best\_answer\_no\_instructions\_"\\
\midrule    
c9 & 
        "natural\_questions\_open\_1\_0\_0",
        "web\_questions\_whats\_the\_answer",
        "web\_questions\_question\_answer",
        "dbpedia\_14\_pick\_one\_category\_for\_the\_following\_text",
        "kilt\_tasks\_hotpotqa\_combining\_facts",
        "web\_questions\_short\_general\_knowledge\_q",
        "kilt\_tasks\_hotpotqa\_straighforward\_qa",
        "adversarial\_qa\_dbidaf\_generate\_question",
        "adversarial\_qa\_droberta\_based\_on",
        "web\_questions\_get\_the\_answer",
        "kilt\_tasks\_hotpotqa\_complex\_question",
        "web\_questions\_potential\_correct\_answer",
        "trivia\_qa\_rc\_1\_1\_0",
        "kilt\_tasks\_hotpotqa\_formulate",
        "adversarial\_qa\_dbert\_based\_on",
        "adversarial\_qa\_dbidaf\_based\_on",
        "squad\_v1\_1\_3\_0\_0" \\
\bottomrule

\end{tabularx}
\caption{Task names for each of the 10 clusters obtained by applying \texttt{MBC} clustering to Phi-2 private library with 256 experts, with each expert trained for 2 epochs.}
  \label{tab:zero_shot_sni_phi3}
\end{table*}

\begin{table*}[htbp]
\centering
\scriptsize
\begin{tabularx}{\textwidth}{|l|X}
\toprule
c0 & 
        "adversarial\_qa\_dbert\_generate\_question", "adversarial\_qa\_dbidaf\_generate\_question", "adversarial\_qa\_droberta\_generate\_question", "app\_reviews\_generate\_review", "cot\_creak", "cot\_esnli", "cot\_esnli\_ii", "dream\_generate\_first\_utterance", "dream\_generate\_last\_utterance", "duorc\_ParaphraseRC\_title\_generation", "duorc\_SelfRC\_title\_generation", "fix\_punct", "gem\_common\_gen\_1\_1\_0", "gem\_dart\_1\_1\_0", "gigaword\_1\_2\_0", "huggingface\_xsum", "lambada\_1\_0\_0", "race\_high\_Write\_a\_multi\_choice\_question\_for\_the\_following\_article", "race\_high\_Write\_a\_multi\_choice\_question\_options\_given\_", "race\_middle\_Write\_a\_multi\_choice\_question\_for\_the\_following\_article", "race\_middle\_Write\_a\_multi\_choice\_question\_options\_given\_", "stream\_aqua", "stream\_qed", "wiqa\_what\_is\_the\_missing\_first\_step", "wmt16\_translate\_fi\_en\_1\_0\_0", "wmt16\_translate\_ro\_en\_1\_0\_0", "yelp\_polarity\_reviews\_0\_2\_0"
        \\
\midrule    
c1 & 
        "ag\_news\_subset\_1\_0\_0", "app\_reviews\_convert\_to\_rating", "app\_reviews\_convert\_to\_star\_rating", "cot\_creak\_ii", "cot\_ecqa\_ii", "cot\_gsm8k\_ii", "cot\_sensemaking\_ii", "cot\_strategyqa", "dbpedia\_14\_given\_a\_choice\_of\_categories\_", "dbpedia\_14\_given\_a\_list\_of\_category\_what\_does\_the\_title\_belong\_to", "dbpedia\_14\_given\_list\_what\_category\_does\_the\_paragraph\_belong\_to", "glue\_mnli\_2\_0\_0", "glue\_qnli\_2\_0\_0", "glue\_qqp\_2\_0\_0", "glue\_stsb\_2\_0\_0", "glue\_wnli\_2\_0\_0", "kilt\_tasks\_hotpotqa\_complex\_question", "paws\_wiki\_1\_1\_0", "qasc\_is\_correct\_1", "qasc\_is\_correct\_2", "snli\_1\_1\_0", "social\_i\_qa\_Check\_if\_a\_random\_answer\_is\_valid\_or\_not", "social\_i\_qa\_Generate\_answer", "social\_i\_qa\_Generate\_the\_question\_from\_the\_answer", "social\_i\_qa\_I\_was\_wondering", "squad\_v1\_1\_3\_0\_0", "squad\_v2\_0\_3\_0\_0", "stream\_qed\_ii", "super\_glue\_multirc\_1\_0\_2", "super\_glue\_rte\_1\_0\_2", "super\_glue\_wic\_1\_0\_2", "super\_glue\_wsc\_fixed\_1\_0\_2", "trec\_1\_0\_0", "wiki\_bio\_guess\_person", "wiki\_qa\_Is\_This\_True\_"
        \\
\midrule    
c2 & 
        "app\_reviews\_categorize\_rating\_using\_review", "cos\_e\_v1\_11\_question\_option\_description\_text", "cot\_qasc", "cot\_strategyqa\_ii", "dbpedia\_14\_pick\_one\_category\_for\_the\_following\_text", "definite\_pronoun\_resolution\_1\_1\_0", "kilt\_tasks\_hotpotqa\_final\_exam", "math\_dataset\_algebra\_\_linear\_1d\_1\_0\_0", "qasc\_qa\_with\_separated\_facts\_4", "quarel\_do\_not\_use", "quoref\_Context\_Contains\_Answer", "race\_high\_Is\_this\_the\_right\_answer", "race\_middle\_Is\_this\_the\_right\_answer", "sciq\_Direct\_Question", "sciq\_Multiple\_Choice", "sciq\_Multiple\_Choice\_Closed\_Book\_", "sciq\_Multiple\_Choice\_Question\_First", "social\_i\_qa\_Show\_choices\_and\_generate\_index", "stream\_aqua\_ii", "super\_glue\_cb\_1\_0\_2", "super\_glue\_copa\_1\_0\_2", "unified\_qa\_science\_inst", "wiki\_qa\_Decide\_good\_answer", "wiki\_qa\_Direct\_Answer\_to\_Question", "wiki\_qa\_Generate\_Question\_from\_Topic", "wiki\_qa\_Jeopardy\_style", "wiki\_qa\_Topic\_Prediction\_Answer\_Only", "wiki\_qa\_Topic\_Prediction\_Question\_Only", "wiki\_qa\_Topic\_Prediction\_Question\_and\_Answer\_Pair", "wiki\_qa\_automatic\_system", "wiki\_qa\_exercise", "wiki\_qa\_found\_on\_google"
        \\
\midrule    
c3 & 
        "adversarial\_qa\_dbert\_answer\_the\_following\_q", "adversarial\_qa\_dbert\_based\_on", "adversarial\_qa\_dbert\_question\_context\_answer", "adversarial\_qa\_dbert\_tell\_what\_it\_is", "adversarial\_qa\_dbidaf\_answer\_the\_following\_q", "adversarial\_qa\_dbidaf\_based\_on", "adversarial\_qa\_dbidaf\_question\_context\_answer", "adversarial\_qa\_dbidaf\_tell\_what\_it\_is", "adversarial\_qa\_droberta\_answer\_the\_following\_q", "adversarial\_qa\_droberta\_based\_on", "adversarial\_qa\_droberta\_question\_context\_answer", "adversarial\_qa\_droberta\_tell\_what\_it\_is", "cos\_e\_v1\_11\_aligned\_with\_common\_sense", "cos\_e\_v1\_11\_explain\_why\_human", "cos\_e\_v1\_11\_generate\_explanation\_given\_text", "cos\_e\_v1\_11\_i\_think", "cos\_e\_v1\_11\_rationale", "drop\_2\_0\_0", "duorc\_ParaphraseRC\_generate\_question\_by\_answer", "duorc\_SelfRC\_generate\_question\_by\_answer", "kilt\_tasks\_hotpotqa\_combining\_facts", "kilt\_tasks\_hotpotqa\_formulate", "kilt\_tasks\_hotpotqa\_straighforward\_qa", "natural\_questions\_open\_1\_0\_0", "trivia\_qa\_rc\_1\_1\_0", "web\_questions\_get\_the\_answer", "web\_questions\_potential\_correct\_answer", "web\_questions\_question\_answer", "web\_questions\_short\_general\_knowledge\_q", "web\_questions\_whats\_the\_answer"
        \\
\midrule    
c4 & 
        "duorc\_ParaphraseRC\_answer\_question", "duorc\_ParaphraseRC\_decide\_worth\_it", "duorc\_ParaphraseRC\_extract\_answer", "duorc\_ParaphraseRC\_generate\_question", "duorc\_ParaphraseRC\_movie\_director", "duorc\_ParaphraseRC\_question\_answering", "duorc\_SelfRC\_answer\_question", "duorc\_SelfRC\_decide\_worth\_it", "duorc\_SelfRC\_extract\_answer", "duorc\_SelfRC\_generate\_question", "duorc\_SelfRC\_movie\_director", "duorc\_SelfRC\_question\_answering", "quac\_1\_0\_0", "quoref\_Answer\_Friend\_Question", "quoref\_Answer\_Test", "quoref\_Find\_Answer", "quoref\_Found\_Context\_Online", "quoref\_Given\_Context\_Answer\_Question", "quoref\_Guess\_Answer", "quoref\_Guess\_Title\_For\_Context", "quoref\_Read\_And\_Extract\_", "quoref\_What\_Is\_The\_Answer"
        \\
\midrule    
c5 & 
        "cos\_e\_v1\_11\_description\_question\_option\_id", "cos\_e\_v1\_11\_question\_description\_option\_id", "dream\_baseline", 
        \newline"dream\_read\_the\_following\_conversation\_and\_answer\_the\_question", "quail\_context\_description\_question\_answer\_id", "quail\_context\_description\_question\_answer\_text", "quail\_context\_description\_question\_text", "quail\_context\_question\_answer\_description\_id", "quail\_context\_question\_answer\_description\_text", "quail\_context\_question\_description\_answer\_id", "quail\_context\_question\_description\_answer\_text", "quail\_context\_question\_description\_text", "quail\_description\_context\_question\_answer\_id", "quail\_description\_context\_question\_answer\_text", "quail\_description\_context\_question\_text", "quail\_no\_prompt\_id", "quail\_no\_prompt\_text", "race\_high\_Read\_the\_article\_and\_answer\_the\_question\_no\_option\_", "race\_high\_Select\_the\_best\_answer", "race\_high\_Select\_the\_best\_answer\_generate\_span\_", "race\_high\_Select\_the\_best\_answer\_no\_instructions\_", "race\_high\_Taking\_a\_test", "race\_middle\_Read\_the\_article\_and\_answer\_the\_question\_no\_option\_", "race\_middle\_Select\_the\_best\_answer", "race\_middle\_Select\_the\_best\_answer\_generate\_span\_", "race\_middle\_Select\_the\_best\_answer\_no\_instructions\_", "race\_middle\_Taking\_a\_test"
        \\
\midrule    
c6 & 
        "cos\_e\_v1\_11\_description\_question\_option\_text", "cos\_e\_v1\_11\_question\_description\_option\_text", "cos\_e\_v1\_11\_question\_option\_description\_id", "qasc\_qa\_with\_combined\_facts\_1", "qasc\_qa\_with\_separated\_facts\_1", "qasc\_qa\_with\_separated\_facts\_2", "qasc\_qa\_with\_separated\_facts\_3", "qasc\_qa\_with\_separated\_facts\_5", "quarel\_choose\_between", "quarel\_heres\_a\_story", "quarel\_logic\_test", "quarel\_testing\_students", "quartz\_answer\_question\_based\_on", "quartz\_answer\_question\_below", "quartz\_given\_the\_fact\_answer\_the\_q", "quartz\_having\_read\_above\_passage", "quartz\_paragraph\_question\_plain\_concat", "quartz\_read\_passage\_below\_choose", "quartz\_use\_info\_from\_paragraph\_question", "quartz\_use\_info\_from\_question\_paragraph", "quoref\_Answer\_Question\_Given\_Context", "ropes\_background\_new\_situation\_answer", "ropes\_background\_situation\_middle", "ropes\_given\_background\_situation", "ropes\_new\_situation\_background\_answer", "ropes\_plain\_background\_situation", "ropes\_plain\_bottom\_hint", "ropes\_plain\_no\_background", "ropes\_prompt\_beginning", "ropes\_prompt\_bottom\_hint\_beginning", "ropes\_prompt\_bottom\_no\_hint", "ropes\_prompt\_mix", "ropes\_read\_background\_situation", "sciq\_Direct\_Question\_Closed\_Book\_", "social\_i\_qa\_Show\_choices\_and\_generate\_answer", "wiqa\_does\_the\_supposed\_perturbation\_have\_an\_effect", "wiqa\_effect\_with\_label\_answer", "wiqa\_effect\_with\_string\_answer", "wiqa\_which\_of\_the\_following\_is\_the\_supposed\_perturbation"
        \\
\midrule    
c7 & 
        "aeslc\_1\_0\_0", "cnn\_dailymail\_3\_4\_0", "coqa\_1\_0\_0", "cot\_gsm8k", "dream\_answer\_to\_dialogue", "duorc\_ParaphraseRC\_build\_story\_around\_qa", "duorc\_SelfRC\_build\_story\_around\_qa", "gem\_e2e\_nlg\_1\_1\_0", "gem\_web\_nlg\_en\_1\_1\_0", "gem\_wiki\_lingua\_english\_en\_1\_1\_0", "multi\_news\_1\_0\_0", "wiki\_bio\_comprehension", "wiki\_bio\_key\_content", "wiki\_bio\_what\_content", "wiki\_bio\_who", "wiqa\_what\_is\_the\_final\_step\_of\_the\_following\_process", "wiqa\_what\_might\_be\_the\_first\_step\_of\_the\_process", "wiqa\_what\_might\_be\_the\_last\_step\_of\_the\_process", "wmt16\_translate\_tr\_en\_1\_0\_0"
        \\
\midrule    
c8 & 
        "anli\_r1\_0\_1\_0", "anli\_r2\_0\_1\_0", "anli\_r3\_0\_1\_0", "cosmos\_qa\_1\_0\_0", "cot\_ecqa", "cot\_sensemaking", "glue\_cola\_2\_0\_0", "glue\_mrpc\_2\_0\_0", "glue\_sst2\_2\_0\_0", "imdb\_reviews\_plain\_text\_1\_0\_0", "para\_crawl\_enes", "super\_glue\_record\_1\_0\_2", "true\_case", "wmt14\_translate\_fr\_en\_1\_0\_0", "wmt16\_translate\_de\_en\_1\_0\_0", "word\_segment"
        \\
\midrule    
c9 &
    "wiki\_hop\_original\_choose\_best\_object\_affirmative\_1", "wiki\_hop\_original\_choose\_best\_object\_affirmative\_2", "wiki\_hop\_original\_choose\_best\_object\_affirmative\_3", "wiki\_hop\_original\_choose\_best\_object\_interrogative\_1", "wiki\_hop\_original\_choose\_best\_object\_interrogative\_2", "wiki\_hop\_original\_explain\_relation", "wiki\_hop\_original\_generate\_object", "wiki\_hop\_original\_generate\_subject", "wiki\_hop\_original\_generate\_subject\_and\_object"
    \\
\bottomrule
\end{tabularx}
\caption{Task names for each of the 10 clusters obtained by applying \texttt{MBC} clustering to Mistral 7B private library with 256 experts.}
  \label{tab:task_cluster_names}
\end{table*}

\end{document}